\documentclass{article}

    \usepackage[final]{neurips_2024}

\usepackage[utf8]{inputenc} % allow utf-8 input
\usepackage[T1]{fontenc}    % use 8-bit T1 fonts
\usepackage{hyperref}       % hyperlinks
\usepackage{url}            % simple URL typesetting
\usepackage{booktabs}       % professional-quality tables
\usepackage{amsfonts}       % blackboard math symbols
\usepackage{nicefrac}       % compact symbols for 1/2, etc.
\usepackage{microtype}      % microtypography
\usepackage[dvipsnames]{xcolor}        % colors
\usepackage{amsmath}
\usepackage{cleveref}
\usepackage{pifont}
\usepackage{enumitem}
\usepackage{multicol}
\usepackage[most]{tcolorbox}

\usepackage[toc,page]{appendix}
\usepackage{multirow}
\usepackage{tablefootnote}
\usepackage{bbm}
\usepackage{wrapfig}
\usepackage[ruled]{algorithm2e}
\usepackage{xspace}
\newcommand{\ourmethod}{{\fontfamily{lmtt}\selectfont \textbf{GDeR}}\xspace}

\definecolor{darkgrey}{RGB}{120,120,120}
\definecolor{mygrey}{RGB}{200,200,200}
\definecolor{RedOrange}{RGB}{255,69,0}
\SetKwInOut{Input}{Input}\SetKwInOut{Output}{Output}
\SetKwComment{Comment}{$\triangleright$\ }{}

\title{\ourmethod: Safeguarding Efficiency, Balancing, and Robustness via Prototypical Graph Pruning}

\hypersetup{
    colorlinks=true,
    linkcolor=red,
    citecolor=green,
    filecolor=magenta,      
    urlcolor=magenta,
    }

\author{%
  \small Guibin Zhang$^{1,2*}$, Haonan Dong$^{1*}$, Yuchen Zhang$^{2}$, Zhixun Li$^{3}$, Dingshuo Chen$^{4}$,  \\ \small \textbf{Kai Wang$^{5}$}, \textbf{Tianlong Chen$^{6}$}, \textbf{Yuxuan Liang$^{7}$}, \textbf{Dawei Cheng$^{\dag 1,2}$}, \textbf{Kun Wang$^{\dag 8}$} \\
  \small$^{1}$Tongji Univerity, $^{2}$Shanghai AI Laboratory, $^{3}$CUHK,  \small$^{4}$UCAS, \\
  \small$^{5}$NUS, $^{6}$UNC-Chapel Hill, $^{7}$HKUST (Guangzhou) $^{8}$NTU \\ \footnotesize	 $^*$ Equal Contribution, $^\dag$ Corresponding author \\
  \small \texttt{dcheng@tongji.edu.cn}, \;\texttt{wk520529wjh@gmail.com} 
}

\begin{document}

\maketitle

\begin{abstract}
\vspace{-0.7em}
Training high-quality deep models necessitates vast amounts of data, resulting in overwhelming computational and memory demands. Recently, data pruning, distillation, and coreset selection have been developed to streamline data volume by \textit{retaining}, \textit{synthesizing}, or \textit{selecting} a small yet informative subset from the full set. Among these methods, data pruning incurs the least additional training cost and offers the most practical acceleration benefits. However, it is the most vulnerable, often suffering significant performance degradation with imbalanced or biased data schema, thus raising concerns about its accuracy and reliability in on-device deployment. Therefore, there is a looming need for a new data pruning paradigm that maintains the efficiency of previous practices while ensuring balance and robustness.
Unlike the fields of computer vision and natural language processing, where mature solutions have been developed to address these issues, graph neural networks (GNNs) continue to struggle with increasingly large-scale, imbalanced, and noisy datasets, lacking a unified dataset pruning solution. 
To achieve this, we introduce a novel dynamic soft-pruning method, \ourmethod, designed to update the training ``basket'' during the process using trainable prototypes. \ourmethod first constructs a well-modeled graph embedding hypersphere and then samples \textit{representative, balanced, and unbiased subsets} from this embedding space, which achieves the goal we called {\fontfamily{lmtt}\selectfont \textbf{Graph Training Debugging}}.
Extensive experiments on five datasets across three GNN backbones, demonstrate that \ourmethod (I) achieves or surpasses the performance of the full dataset with $30\%\sim50\%$ fewer training samples, (II) attains up to a $2.81\times$ lossless training speedup, and (III) outperforms state-of-the-art pruning methods in imbalanced training and noisy training scenarios by $0.3\%\sim4.3\%$ and $3.6\%\sim7.8\%$, respectively. The source code is available at \texttt{\url{https://github.com/ins1stenc3/GDeR}}.
\vspace{-0.5em}

\end{abstract}
\vspace{-0.7em}
\section{Introduction}\label{sec:intro}
\vspace{-0.7em}
Data-centric AI, though continuously providing high-quality data for upcoming artificial general intelligence \cite{motamedi2021data, zha2023data, floridi2020gpt, li2024camel},  presents a significant hurdle for their on-device deployment during training and inference phases \cite{frantar2023sparsegpt, ashkboos2024slicegpt, sun2023simple, sanh2020movement}. To democratize existing state-of-the-art methods \cite{touvron2023llama, achiam2023gpt, touvron2023llama, wang2023chatvideo, chen2023minigptv2, zhu2023minigpt, chen2023vast}, considerable efforts are directed toward identifying unbiased and core data within training datasets and conducting troubleshooting to deepen our solid understanding of the intrinsic property of data shcema  \cite{karlaš2022data, shim2021core, tukan2020coresets, paul2021deep}. 
To date, \textit{data pruning} \cite{zhang2024graph, wang2023brave, zhang2024heads, raju2021accelerating, qin2023infobatch}, \textit{distillation} \cite{zhao2023dataset, wang2022cafe, cazenavette2022dataset, nguyen2021dataset, li2023attend, zhangnavigating, zhang2024two2} and \textit{coreset selection} \cite{har2004coresets, chen2009coresets, toneva2018empirical, shim2021core} aim to retain, synthesize or choose a small but informative dataset from original full set. While the sample size undergoes significant reshaping and reduction, methods like dataset distillation inevitably lead to additional training costs \cite{qin2023infobatch,zhao2020dataset, lu2023can}. As a hardware-friendly candidate and accelerator for training and inference, data pruning serves as a promising candidate by mitigating the high computational burden.

\begin{wrapfigure}{r}{0.5\textwidth}
\vspace{-0.5em}
  \begin{center}
    \includegraphics[width=0.5\textwidth]{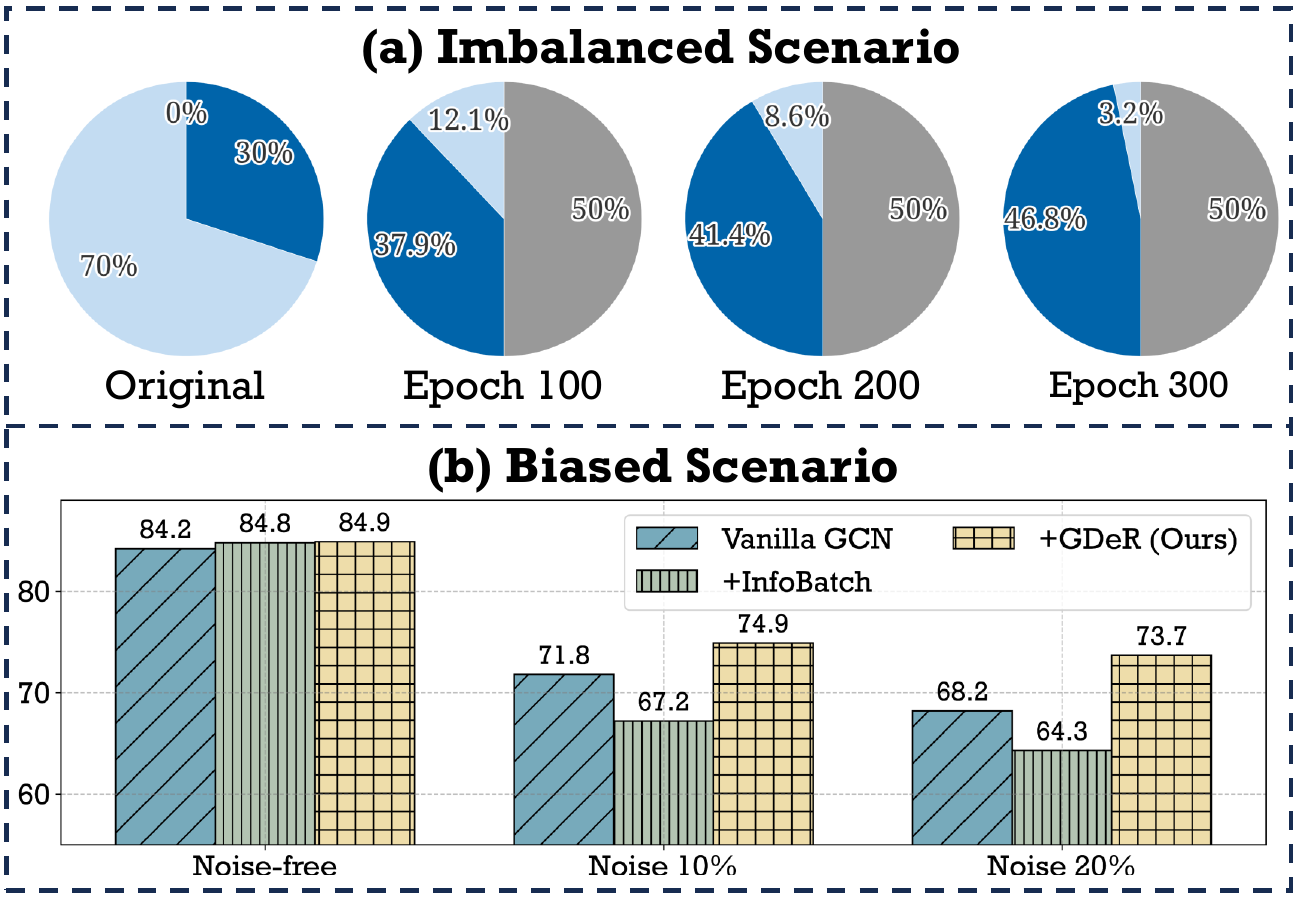}
  \end{center}
  \vspace{-1em}
  \caption{({\textbf{a}}) We report the label distribution of the training set retained by InfoBatch at pruning ratios of $50\%$ in the $\{0,100,200,300\}$-th epochs. The gray, light blue and dark blue represent pruned, minority, and majority samples, respectively. ({\textbf{b}}) Performance comparison between InfoBatch and our \ourmethod when introducing outliers (following \cite{li2022graphde}) into $\{0\%,10\%,20\%\}$ of the training set.}\label{fig:intro}
  \vspace{-1.2em}
\end{wrapfigure}

Recent advancements, however, have demonstrated that the efficiency of data pruning may come at a cost—utilizing only a portion of the data can potentially render the model more vulnerable to imbalance or malicious perturbation attacks~\cite{park2024robust}, which are commonly seen in real-world applications. As illustrated in \Cref{fig:intro}, we evaluate the performance of the current state-of-the-art data pruning paradigm, InfoBatch~\cite{qin2023infobatch}, with a pruning ratio of $50\%$, on \textsc{mutag}~\cite{debnath1991structure}+GCN~\cite{kipf2016} under both imbalance and biased scenarios. It can be observed that: \ding{182} InfoBatch exacerbates the imbalance of training samples during the training process; \ding{183} InfoBatch efficiently saves training costs in noise-free scenarios, even surpassing the original full dataset performance by $0.7\%$. However, it encounters significant performance degradation ($5.7\%\sim6.4\%\downarrow$) in biased scenarios. 

In this paper, we primarily focus on \textit{graph-level data pruning}, aiming to enhance the model's robustness to data imbalance and noise while maintaining the efficiency inherited from traditional data pruning practices. This is because, unlike in computer vision (CV) and natural language processing (NLP) domains, where separate solutions already exist for addressing these issues~\cite{park2024robust,qin2023infobatch}, graph learning models continue to grapple with increasingly large-scale, imbalanced, and biased datasets~\cite{wang2022imbalanced,li2022graphde,li2024rethinking,S3GCL_ICML24,FGGP_AAAI24,FedSSP_NeurIPS,FGSSL_IJCAI23}. To this end, we first introduce a novel direction in the realm of graph training, termed {\fontfamily{lmtt}\selectfont \textbf{Graph Training Debugging (GTD)}}, to (dynamically) identify \textit{representative, robust, and unbiased subsets} for accelerating the training process without compromising performance. 

\vspace{-0.1em}
We achieve  {\fontfamily{lmtt}\selectfont \textbf{GTD}} goal by proposing a novel dynamical soft-pruning method, {\fontfamily{lmtt}\selectfont \textbf{Graph De-Redundancy (GDeR)}}, in which specifically designed to work efficiently and accurately on various GNN architectures. Concretely, \ourmethod draws inspiration from prototype learning~\cite{arik2020protoattend,li2020prototypical} practices, projecting training graph samples onto a hyperspherical embedding space. It utilizes a set of trainable prototypes to regularize the graph embedding distribution, essentially encouraging both inter-class separateness and intra-class compactness. Furthermore, on this well-regularized hypersphere, \ourmethod generates a sampling distribution that encourages the sampling of under-learned graphs, while excluding those with high outlier risk and belonging to majority clusters. Given a training budget (\textit{i.e.}, pruning ratio), \ourmethod dynamically maintains a sub-dataset at each epoch, efficiently combating the negative impact of imbalanced and noisy data on the model, simultaneously accelerating training significantly.

\vspace{-1em}
\paragraph{Broader Impact.} In this paper, we present a novel training philosophy {\fontfamily{lmtt}\selectfont \textbf{GDeR}} to achieve our defined {\fontfamily{lmtt}\selectfont \textbf{GTD}} goal. \ourmethod dynamically prunes irrelevant graph samples, providing a more comprehensive insight and achieving a triple-win of \underline{\textit{efficiency}}, \underline{{\textit{balancing}}}, and \underline{{\textit{robustness}}}. This approach can contribute to a wide range of graph-related applications, accelerating model training while demonstrating great potential in scenarios such as adversarial attacks \cite{hoffman2018cycada, tsai2018learning, tsai2018learning}, imbalanced graph classification \cite{wang2022imbalanced, pan2013graph}, and unsupervised pre-training \cite{hou2022graphmae, you2020graph, you2021graph}. We believe GDeR can serve as a benchmark for future research in this area, attracting significant attention and inspiring further exploration into understanding sparsity in other domains such as LLMs.

\vspace{-0.9em}
\paragraph{Experimental Observation} We validate the \ourmethod through a comprehensive series of graph-level tasks, across five datasets and three GNN backbones, showcasing that \ourmethod can: 
\ding{182} achieve lossless training performances with $30\%\sim50\%$ fewer training samples,
\ding{183} achieve a $2.0\times$ lossless speedup on \textsc{ogbg-molhiv}, and a $2.81\times$ lossless speedup on pre-training \textsc{ZINC}.
\ding{184} mitigate imbalance issues by achieving a $0.3\sim4.3\% \uparrow$ in F1-macro on \textsc{mutag} and \textsc{dhfr} datasets,
\ding{185} effectively help outlier-attacked GNNs improve accuracy by $3.5\%\sim10.2\%$ through data pruning.

\vspace{-1.0em}
\paragraph{Limitations \& Future Insight.}  \ourmethod, as a plug-in to graph training, not only improves efficiency but also ensures robustness and balance throughout the training. However, the applicability of its principles in fields such as CV remains unexplored, limiting the generalizability of data debugging. This represents a direction for future development in our work.

% \clearpage
\vspace{-0.6em}
\section{Technical Background}
\vspace{-0.6em}

\paragraph{Notations} 
% \vspace{-0.5em}
Consider an undirected graph $\mathcal{G} = (\mathcal{V}, \mathcal{E})$, where $\mathcal{V}$ represents the node set and $\mathcal{E}$ signifies the edges. The feature matrix for the graph is designated as $\mathbf{X} \in \mathbb{R}^{|\mathcal{V}| \times F}$. Each node $v_i \in \mathcal{V}$ is associated with a feature vector of $F$ dimensions. The adjacency matrix $\mathbf{A} \in \{0,1\}^{N \times N}$ represents the connectivity between nodes, where $\mathbf{A}[i,j] = 1$ suggests the presence of an edge $e_{ij} \in \mathcal{E}$, and 0 indicates no edge. In graph-level training tasks, specifically for graph classification, given a set of $N$ graphs $\{\mathcal{G}\}= \{\mathcal{G}_1, \mathcal{G}_2, \ldots, \mathcal{G}_N\}$, where each graph $\mathcal{G}_i = (\mathcal{V}^{i}, \mathcal{E}^{i})$ is as defined above, and their corresponding labels $\mathbf{Y} \in \mathbb{R}^{N \times C}$ with $C$ being the total number of classes, we aim to learn graph representations $\mathbf{H} \in \mathbb{R}^{N \times d'}$ with $\mathbf{H}[i,:]$ for each $\mathcal{G}_i \in \mathcal{G}$ that effectively predict $\mathbf{Y}_i$.

% \vspace{-0.3em}
\vspace{-0.8em}
\paragraph{Graph Neural Networks (GNNs).}
% \vspace{-0.5em}
GNNs~\cite{wu2020comprehensive, zhou2020graph} have become pivotal for learning graph representations, achieving benchmark performances in various graph tasks at \textit{node-level}~\cite{xiao2022graph}, \textit{edge-level}~\cite{zhang2018link}, and \textit{graph-level}~\cite{liu2022graph}. The success of GNN mainly stems from message-passing mechanism:
\vspace{-0.3em}
\begin{equation}\label{eq:gnn}
\mathbf{h}_i^{(l)} = \text{\fontfamily{lmtt}\selectfont \textbf{COMB}}\left( \mathbf{h}_i^{(l-1)}, \text{\fontfamily{lmtt}\selectfont \textbf{AGGR}}\{  \mathbf{h}_j^{(l-1)}: v_j \in \mathcal{N}(v_i) \} \right),\;0\leq l \leq L.
% \vspace{-0.2em}
\end{equation}
Here, $L$ represents the number of GNN layers, where $\mathbf{h}_i^{(0)} = \mathbf{x}_i$, and $\mathbf{h}_i^{(l)} (1\leq l\leq L)$ denotes the node embedding of $v_i$ at the $l$-th layer.  $\mathcal{N}(v_i)$ denotes the 1-hop neighbors of $v_i$, and \text{\fontfamily{lmtt}\selectfont \textbf{AGGR}}($\cdot$) and \text{\fontfamily{lmtt}\selectfont \textbf{COMB}}($\cdot$) are used for aggregating neighborhood information and combining ego/neighbor-representations, respectively. Finally, a sum/mean pooling operation is commonly used for \text{\fontfamily{lmtt}\selectfont \textbf{READOUT}} function to obtain the graph-level embedding.
While promising, the increasing volume of graph samples~\cite{wang2022searching,zhang2024graph,chen2024uncovering} poses significant computational challenges for both training and pre-training of GNNs. Efficiently accelerating graph-level training remains an unresolved issue.

\vspace{-0.9em}
\paragraph{Data Pruning} Current data pruning methods can be categorized as static or dynamic~\cite{qin2023infobatch}. Static data pruning involves heuristic-based metrics or limited training to assess sample importance and perform pruning before formal training, like EL2N~\cite{paul2021deep} and Influence-score~\cite{koh2017understanding}. On the other hand, dynamic data pruning dynamically selects different training samples during training~\cite{raju2021ddp,qin2023infobatch,chen2024beyond}, often achieving better results than static pruning. In the graph domain, attempts related to data pruning include edge-level sampling techniques like GraphSAGE~\cite{sui2021inductive} and GraphSAINT~\cite{zeng2019graphsaint}. However, to the best of our knowledge, there is currently no method specially designed for graph-level data pruning, let alone one that can simultaneously improve the balance and robustness of GNNs.

\vspace{-0.9em}
\paragraph{Imbalance in GNNs} Deep imbalanced learning has been one of the significant challenges in deep learning~\cite{zhang2023deep}. The current mainstream research can be broadly categorized into three approaches: (1) \textit{re-sampling}~\cite{he2009learning,chawla2002smote,kang2019decoupling}, which balances the number of samples from different classes; (2) \textit{re-balancing}~\cite{lin2017focal,cui2019class,tan2020equalization}, which adjusts the loss values for samples from different classes; and (3) \textit{post-hoc processing}~\cite{menon2020long}, which shifts the model logits based on label frequencies. In the domain of graph learning, most efforts to address the imbalance issue focus on node-level classification imbalances~\cite{DRGCN,graphsmote,qu2021imgagn}, yet solutions targeting graph-level imbalance are relatively limited. Despite a preliminary attempt~\cite{wang2022imbalanced}, which requires complex up-sampling and regrouping operations, there is still a need for a straightforward yet effective solution to graph imbalance issue.
% \vspace{-0.5em}

\vspace{-0.8em}
\paragraph{Robustness in GNNs}
% \vspace{-0.5em}

As for robustness learning, many studies showcase graph classification is vulnerable to adversarial
attacks \cite{ma2020towards, zhu2019robust}. Given a set of training or test graphs, an attacker could perturb the graph structure \cite{li2024gslb} and/or node features to deceive a graph classifier into making incorrect predictions for the perturbed testing graph. Traditional empirical and certified defenses \cite{chen2017practical, wang2023turning, yang2023graphguard, zhang2021backdoor} often involve complex designs and additional components. In this paper, we propose subtle adjustments during training, leveraging prototypes to enhance the robustness of graph training.

\vspace{-0.5em}
\section{Methodology}
\vspace{-0.5em}
\subsection{Problem Formulation}
\vspace{-0.7em}
In the classic scenario of graph-level training (not limited to specific tasks like graph classification, regression, or pre-training), given a graph dataset $\mathcal{D} = \{z_i\}_{i=1}^{|\mathcal{D}|} = \{(\mathcal{G}_i, \mathbf{Y}_i)\}_{i=1}^{|\mathcal{D}|}$, a GNN encoder is employed to extract graph-level embeddings $\mathbf{H} = \{\mathbf{h}_i\}_{i=1}^{|\mathcal{D}|}$ for each graph sample, which are then utilized for downstream tasks. The goal of \ourmethod is to find an oracle function that changes with time (epochs) and can determine the current most representative, balanced, and denoised core subset $\mathcal{X}_t$:
\begin{equation}\label{eq:formulation}
\mathcal{X}_{t} = \mathcal{F}_{t-1}\left(\mathcal{D},\{\mathbf{h}^{(t-1)}_i\}_{i=1}^{|\mathcal{D}|}\right),
\end{equation}
where $\mathcal{F}_{t-1}$ is the selection function at the $(t-1)$-th epoch. Given a preset sparsity ratio $s\%$, the subset's volume is fixed as $|\mathcal{X}_{t}| = (1-s)\% \times |\mathcal{D}|$.

\begin{figure}[!t]
\setlength{\abovecaptionskip}{6pt}
\centering
\includegraphics[width=\textwidth]{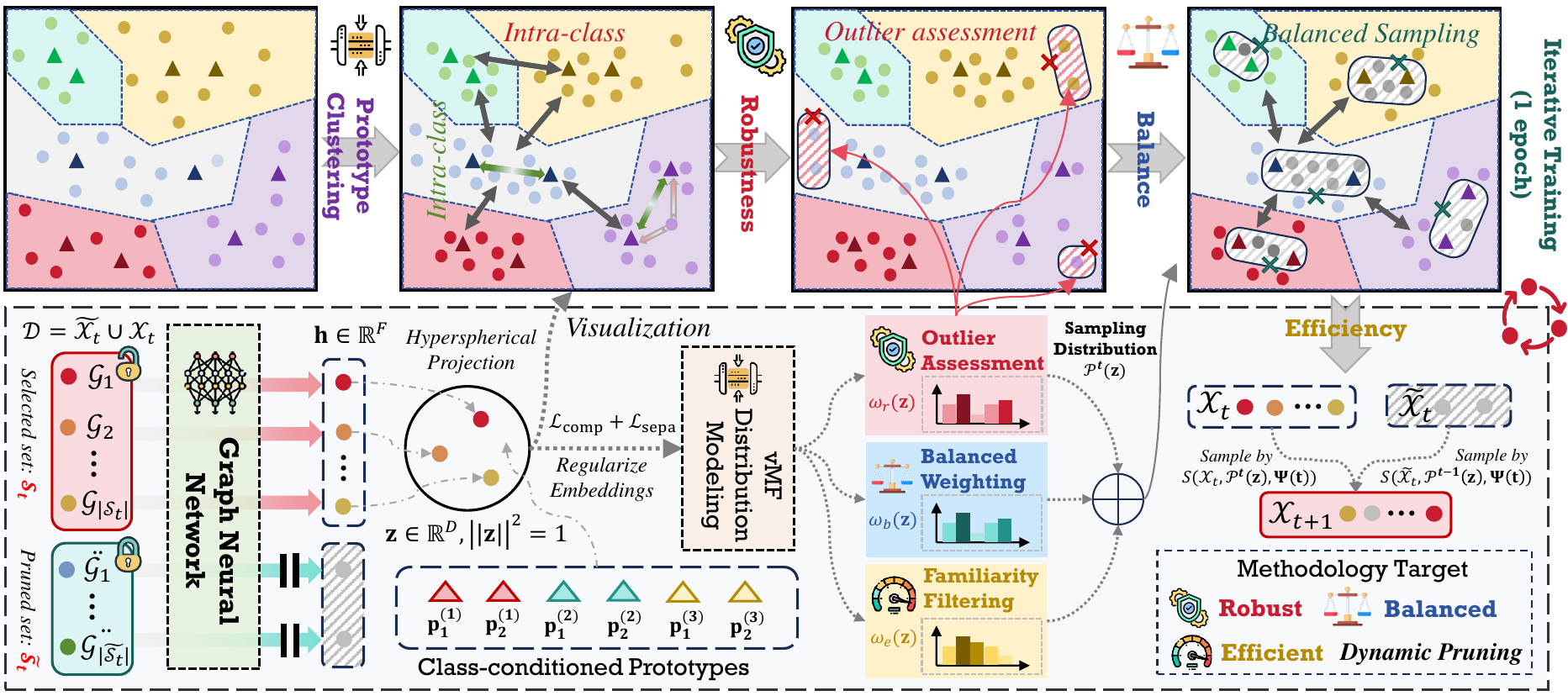}
\vspace{-1.4em}
\caption{The overview of our proposed \ourmethod. \ourmethod comprises hypersphere projection, embedding space modeling, sampling distribution formatting, and the final dynamic sampling. We present the dynamic sample selection process of \ourmethod within one epoch.} \label{fig:framework}
\vspace{-0.9em}
\end{figure}

\subsection{Overview of the Proposed Method}
\vspace{-0.7em}
As shown in \Cref{fig:framework}, given an arbitrary GNN, \ourmethod selects a training sample set $\mathcal{X}_t$ within a specified budget for each epoch. At the $t$-th epoch, after the GNN $f_\theta:\; \mathbf{X}\rightarrow \mathbb{R}^E$ outputs graph embeddings $\mathbf{h}\in\mathbb{R}^E$ from the input graph $\mathcal{G}_i$ with $\mathbf{h} = f_\theta(\mathcal{G}_i)$, these are projected into a hyperspherical embedding space via a \textit{projector} $g_\phi:\;\mathbb{R}^E\rightarrow\mathbb{R}^D$. \ourmethod allocates a set of $M$ trainable prototypes $\mathbf{P}^c = \{\mathbf{p}_k^c\}_{k=1}^K$ for each class $c$, with associated losses used to shape the embedding space, ensuring inter-class separation and intra-class compactness. In this regularized space, \ourmethod formulates a sampling distribution by focusing on samples unfamiliar to the model, excluding those from the majority prototype cluster and with high outlier risk, thereby providing a subset of samples \(S_{t+1}\) for the next epoch. Through this balanced and robust dynamic pruning mechanism, \ourmethod achieves unbiased graph representations at a significantly lower training cost than the full dataset.

\vspace{-0.4em}
\subsection{Projection onto Hyperspherical Embedding Space}
\vspace{-0.5em}
At the $t$-th epoch, \ourmethod maintains a subset $\mathcal{X}_t$ with a given budget, where $s\% = |\mathcal{X}_t| / |\mathcal{D}|$ is a constant, representing the dataset pruning ratio. Given the feature representations $\mathbf{H} \in \mathbb{R}^{|\mathcal{X}_t| \times E}$ output by $f_\theta$, we first project these features into a hyperspherical embedding space, denoted as $\mathbf{z}' = g_\phi(\mathbf{h}), \mathbf{z} = \mathbf{z}'/||\mathbf{z}'||_2$. This projection has been shown to be beneficial for compactly embedding samples of the same class~\cite{wang2020understanding,khosla2020supervised,lu2024learning}. The projected embeddings $\mathbf{z} \in \mathbb{R}^D$, which lie on the unit sphere ($||\mathbf{z}||^2 = 1$), can naturally be modeled using the von Mises-Fisher (vMF) distribution~\cite{wang2020understanding,khosla2020supervised}. Here, we first consider the graph classification scenario\footnote{The extension of \ourmethod to broader scenarios will be detailed in \Cref{sec:opt}}, in which we allocate $K$ prototypes $\mathbf{P}^c = \{\mathbf{p}_k^c\}_{k=1}^K$ for each class $c \;(1\leq c\leq C)$. Following conventional practices in hyperspherical space modeling~\cite{du2022siren}, we model a vMF distribution as the combination of a center prototype representation $\mathbf{p}_k$ and the concentration
parameter $\kappa$:
\begin{equation}\label{eq:distribution}
p_D(\mathbf{z};\mathbf{p}_k,\kappa) = Z_D(\kappa)\exp\left(\kappa \mathbf{p}_k^\top \mathbf{z}\right),\;Z_D(\kappa)=\frac{\kappa^{D/2-1}}{(2\pi)^{D/2}I_{D/2-1}(\kappa)},
\end{equation}
where $\kappa \geq 0$ denotes the tightness around the mean, $Z_D(\kappa)$ represents a normalization factor~\cite{du2022siren}, $\exp\left(\kappa \mathbf{p}_k^\top \mathbf{z}\right)$ is called the angular distance and $I_v$ is the modified Bessel function of the first kind with order $v$. In our multi-prototype settings, we model the probability density of a graph embedding $\mathbf{z}_i$ in class $c$ as follows:
\begin{equation}
p(\mathbf{z}_i;\mathbf{P}^c,\kappa) = \sum_{k=1}^K Z_D(\kappa)\exp(\kappa {\mathbf{p}_k^c}^\top \mathbf{z}_i), 
\end{equation}
%\operatorname{sim}(\mathbf{z}_i, \mathbf{p}_k^c)
Further, the embedding $\mathbf{z}_i$ is assigned to class $c$ with the normalized probability as shown above:
\begin{equation}\label{eq:mixture_vmf}
p(y_i = c\;|\;\mathbf{z}_i;\;\{\mathbf{P}^j,\kappa\}_{j=1}^C) = \frac{\sum_{k=1}^KZ_D(\kappa)\exp({\mathbf{p}_k^c}^\top \mathbf{z}_i/\tau)}{\sum_{j=1}^C\sum_{k'=1}^KZ_D(\kappa')\exp({\mathbf{p}_{k'}^j}^\top \mathbf{z}_i/\tau)},
\end{equation}
where $\tau$ is a temperature coefficient. 
% The class index that $\mathbf{z}_i$ belongs to can be expressed as $\Tilde{y}_i = \operatorname{arg max}_c p(y_i=c)$. 
Given that we have now allocated a corresponding class for each graph embedding, we aim to further encourage: \ding{182} \textit{allocation correctness}, meaning that the allocation should be consistent with the ground truth label; \ding{183} \textit{intra-class compactness}, meaning that graph embeddings should be close to the appropriate prototypes belonging to their own class; and \ding{184} \textit{inter-class separateness}, meaning that graph embeddings should be distant from prototypes of other classes. To achieve \ding{182} and \ding{183} , we have designed the \textit{compactness loss} below:
\begin{equation}\label{eq:loss_gder}
\mathcal{L}_{\text{comp}} = -\frac{1}{|\mathcal{X}_t|}\sum_{i=1}^{|\mathcal{X}_t|}\log \frac{ \sum_{k=1}^K Z_D(\kappa)\exp({\mathbf{p}_k^{y_i}}^\top \mathbf{z}_i / \tau) }{ \sum_{c=1}^C\sum_{k'=1}^K Z_D(\kappa')\exp({\mathbf{p}_{k'}^{y_i}}^\top \mathbf{z}_i / \tau) },
\end{equation}
where $y_i$ represents the class index for $\mathbf{z}_i$. \Cref{eq:loss_gder} is the maximum likelihood estimation of $\max_{\theta,\phi} \Pi_{i=1}^{|\mathcal{X}_t|}p(y_i=c|\mathbf{z}_i,\{\{\mathbf{p}_k^c,\kappa\}_{k=1}^{K}\}_{j=1}^C)$, which not only boosts the allocation correctness but also enforces graph embeddings to compactly surround the appropriate prototypes. Furthermore, to achieve \ding{184}, namely encouraging inter-class separateness, we design the \textit{separation loss}, optimizing large angular  distances among
different class prototypes:
\begin{equation}\label{eq:sep_loss}
\mathcal{L}_{\text{sepa}} = \frac{1}{C}\sum_{i=1}^C\log \frac{1}{C-1} \sum_{j=1}^C \mathbbm{1}_{j\neq i}\sum_{k=1}^{K}\exp (\mathbf{p}_k^j\mathbf{z}_i/\tau)
\end{equation}
where $\mathbbm{1}(\cdot)$ is an indicator function. Through the above regularization, we obtain \(C\) prototype clusters \(\{\chi_c\}_{c=1}^C\), each composed of \(K\) prototype centers \(\{\mathbf{p}_k\}_{k=1}^K\) and surrounding sample sets \(\{\mathbf{z}^{(C)}\}\). After modeling this hypersphere, we proceed with sample selection on the current subset $\mathcal{D}^{(t)}$.

\vspace{-0.5em}
\subsection{Efficient, Balanced and Robust Graph Debugging} 
\vspace{-0.5em}
Traditional dynamic dataset pruning methods typically rely on loss-based metrics to select informative subsets~\cite{raju2021ddp,qin2023infobatch}, which, however, can make the model more vulnerable to imbalance and malicious perturbation (as discussed in \Cref{sec:intro}). In this subsection, while selecting a representative subset $\mathcal{D}^{(t)}$, we also intend to further ensure it is balanced and noise-free. Our first step is to locate samples that are at risk of being outliers in the embedding space. 
%This is because, in real-world scenarios, graph datasets inevitably suffer from noise, perturbations, or outliers~\cite{moonesinghe2008outrank}, which can be detrimental to the GNN training process and could even worsen in the context of dataset pruning. 
We propose using a \textit{prototype-based Mahalanobis distance} to estimate the outlier risk of each graph sample:
\begin{equation}\label{eq:outlier}
\omega_{\text{r}}(\mathbf{z}_i) = -\min\limits_c \left[ -\mathbbm{1}_{y_i \neq c} \max\limits_k \left[ (\mathbf{z}_i - \mathbf{p}_k^c)^\top \Sigma_k^{-1} (\mathbf{z}_i - \mathbf{p}_k^c) \right] \right],\end{equation}
where $\Sigma_k\in\mathbb{R}^{K\times K}$ is the sample covariance of all the prototypes in class $c$. \Cref{eq:outlier} calculates the maximum distance of $\mathbf{z}_i$ to all prototypes within its class, which serves as a robust outlier detection metric~\cite{sehwag2021ssd}. Furthermore, we intend to evaluate the effectiveness of each sample. Given that the distance of an embedding from its cluster center has been shown to be a good indicator of the model's familiarity with it~\cite{abbas2024effective}, we compute the distance of each graph sample to its class-specific prototypes as a familiarity metric:
\begin{equation}\label{eq:familarity}
\omega_{\text{e}}(\mathbf{z}_i) = \frac{\sum_{k=1}^K\operatorname{dist}(\mathbf{p}_k^{y_i},\mathbf{z}_i)}{\sum_{c=1}^C\sum_{k'=1}^K  \mathbbm{1}_{c\neq y_i} \operatorname{dist}(\mathbf{p}_{k'}^{y_i},\mathbf{z}_i)},
\end{equation}
which suggests that if a graph sample is significantly closer to its own prototypes and farther from those of other classes, the model is more familiar with it. We implement the distance function using the angular distance in \Cref{eq:distribution}. When considering the data balancing issue, we formulate the balancing score for each sample $\mathbf{z}_i$ as follows: 
\begin{equation}\label{eq:balance}
\omega_{\text{b}}(\mathbf{z}_i) = {\left|\left\{ \mathbf{z}_i| \min\limits_k \operatorname{dist}(\phi_{\mathbf{z}_i}(\mathbf{p}_k), \mathbf{z}_i)\right\}\right|}/{\left|\{\phi_{{\mathbf{z}_i}}(\chi)\}\right|},
\end{equation}
where $\phi_{\mathbf{z}_i}(\mathbf{p}_k)$ denotes the closest prototype to $\mathbf{z}_i$, and $\phi_{{\mathbf{z}_i}}(\chi)$ denotes the prototype cluster that $\mathbf{z}_i$ currently belongs to. \Cref{eq:balance} evaluates whether the graph sample \(\mathbf{z}_i\) belongs to a minority from a prototype-cluster perspective.
Finally, we assign sampling probabilities to all samples in $\mathcal{D}^{(t)}$:
% \mathcal{P}_{\ourmethod}(\mathbf{z}) = \int_{\mathbf{z}\in \mathcal{D}^{(t)}} \frac{\omega(\mathbf{z})  }{  \int_{\mathbf{z}} \omega(\mathbf{z}) d\mathbf{z}}d\mathbf{z},\; 
\vspace{-0.6em}
\begin{equation}\label{eq:final_distri}
% \vspace{-0.6em}
   \omega(\mathbf{z}_i) = \frac{{\omega_{\text{e}}^\sigma(\mathbf{z}_i)}}{ \left(\omega_{\text{r}}^{\sigma}(\mathbf{z}_i)+\epsilon\right)\cdot \left(\omega_{\text{b}}^\sigma(\mathbf{z}_i)+\epsilon\right)}, 
   %\omega_{\text{b}}(\mathbf{z}_i) = \frac{\left|\left\{ \mathbf{z}_i| \min\limits_k \operatorname{dist}(\mathbf{p}^{y_i}_k, \mathbf{z}_i)\right\}\right|}{\left|\{\chi_{y_i}\}\right|},
\end{equation}
where $(\cdot)^\sigma$ represents the Sigmoid transformation. \Cref{eq:final_distri} is designed to sample with higher probability those samples that the model is less familiar with, have a lower outlier risk, and belong to a minority group. Now, at the $t$-th epoch, we obtain the final sampling probability distribution $\mathcal{P}^{(t)}(\mathbf{z}):\;\int_{\mathbf{z} \in \mathcal{X}_t} \frac{\omega(\mathbf{z})}{\int_{\mathbf{z}} \omega(\mathbf{z}) \, d\mathbf{z}} \, d\mathbf{z}$. Recall that we have $(1-s)\%$ of samples pruned in the $t$-th epoch, \textit{i.e.}, $\Tilde{\mathcal{X}}_{t} = \mathcal{D} \setminus \mathcal{X}_t$. 
For $\Tilde{\mathcal{X}}_{t}$, we use the probability distribution $\mathcal{P}^{(t-1)}(\mathbf{z})$ from the $(t-1)$-th epoch~\footnote{For the first epoch, we set $\mathcal{P}^{(t-1)}(\mathbf{z})$ as 
uniform distribution.}. Specifically, we formulate \ourmethod's coreset sampling function $\mathcal{F}_t$ in \Cref{eq:formulation} as follows:
\begin{equation}
\mathcal{F}_t(\mathcal{D}, \mathbf{H}) = S\left(\mathcal{X}_t, \mathcal{P}^{(t)}(\mathbf{z}),  \Psi(t)\right) \bigcup S\left(\Tilde{\mathcal{X}}_{t}, \mathcal{P}^{(t-1)}(\mathbf{z}), \Tilde{\Psi}(t) \right),
\end{equation}
where $\mathcal{F}_t$ outputs the selected samples $\mathcal{X}_{t+1}$ for the next epoch's training, $S(\mathcal{X},\mathcal{P},N)$ is a sampling operator that samples $N$ samples from $\mathcal{X}$ with probability distribution $\mathcal{P}$, and ${\Psi}(t)$  ($\Tilde{\Psi}(t)$) is the scheduler function (with implementation placed in \Cref{app:scheduler}) that control the number of samples drawn from $\mathcal{X}_{t}$ ($\Tilde{\mathcal{X}}_{t}$), respectively, subject to the given budget ${\Psi}(t) + \Tilde{\Psi}(t) = |\mathcal{D}| \times s\% = |\mathcal{X}_t|$.

\vspace{-0.5em}
\subsection{Optimization and Extension}\label{sec:opt}
\paragraph{Optimization} Aside from the original task-specific loss of GNN training denoted as $\mathcal{L}_{\text{task}}$, \ourmethod has additionally introduced $\mathcal{L}_{\text{comp}}$ and $\mathcal{L}_{\text{sepa}}$. The overall training objective of \ourmethod is formulated as:
\begin{equation}\label{eq:loss}
\mathcal{L}_{\ourmethod} = \mathcal{L}_{\text{task}} + \lambda_1\cdot\mathcal{L}_{\text{comp}} + \lambda_2\cdot\mathcal{L}_{\text{sepa}},
\end{equation}
where $\lambda_1$ and $\lambda_2$ are co-efficient adjusting the relative importance of two losses. 
We conclude the {algorithm workflow} table of \ourmethod in \Cref{app:algo}.

\vspace{-0.7em}
\paragraph{Extension} Finally, we advocate that \ourmethod is not limited to graph classification but can also be seamlessly adapted to tasks such as graph regression and graph pre-training. The key distinction between these tasks and graph classification is that each graph sample does not have a ground truth class index, which makes ground truth class-based calculations, such as those in \Cref{eq:loss_gder,eq:sep_loss}, infeasible. 
One straightforward approach is to manually set $M$ virtual classes, using the class assigned by \Cref{eq:mixture_vmf} as the graph sample’s current class. However, this may result in prototypes and hyperspherical embeddings that do not accurately reflect the underlying clustering distribution~\cite{frigui2004unsupervised}. To address this, we leverage ProtNCELoss~\cite{li2020prototypical} as a self-supervised signal, providing a more reliable reflection of the data’s structure. Detailed implementation can be found in \Cref{app:extend}.

\vspace{-0.5em}
\section{Experiments}
\vspace{-0.9em}
In this section, we conduct extensive experiments to answer the following research questions: 
(\textbf{RQ1}) Can \ourmethod effectively boost GNN efficiency (under both supervised and unsupervised settings)?
(\textbf{RQ2})  Does \ourmethod genuinely accelerate the GNN training?
(\textbf{RQ3}) Can \ourmethod help alleviate graph imbalance?
(\textbf{RQ4}) Can \ourmethod aid in robust GNN training?

\vspace{-0.5em}
\subsection{Experiment Setup}
\vspace{-0.6em}
\paragraph{Datasets and Backbones} We test \ourmethod on two widely-used  datasets, \textsc{mutag}~\cite{debnath1991structure} and \textsc{dhfr}~\cite{sutherland2003spline}; two OGB large-scale datasets, \textsc{ogbg-molhiv} and \textsc{ogbg-molpbca}~\cite{hu2020open}; one large-scale chemical compound dataset \textsc{ZINC}~\cite{irwin2012zinc}. Following \cite{wang2022imbalanced}, we adopt a 25\%/25\%/50\% train/validation/test random split for the \textsc{mutag} and \textsc{dhfr} under imbalanced scenarios and 80\%/10\%/10\% under normal and biased scenarios, both reporting results across 20 data splits. For \textsc{ogbg-molhiv} and \textsc{ogbg-molpbca}, we use the official splits provided by \cite{hu2020open}. For \textsc{ZINC}, we follow the splits specified in \cite{rampavsek2022recipe}. 
% \vspace{-0.7em}
% \paragraph{Backbones and Baselines}
We choose three representative GNNs, including GCN~\cite{kipf2016semi}, PNA~\cite{corso2020principal} and GraphGPS~\cite{rampavsek2022recipe}. Detailed dataset and backbone settings are in \Cref{app:exp_dataset,app:exp_backbone}.
\vspace{-0.8em}
\paragraph{Parameter Configurations} The hyperparameters in \ourmethod include the temperature coefficient $\tau$, prototype count $K$, loss-specific coefficient $\lambda_1$ and $\lambda_2$. Practically, we uniformly set $K=2$, and tune the other three by grid searching: $\tau\in\{1e-3,1e-4,1e-5\}$, $\lambda\in\{1e-1,5e-1\}$,$\lambda\in\{1e-1,1e-5\}$. Detailed ablation study on hyperparameters is placed in \Cref{sec:ablation_sensi}. 

\vspace{-0.5em}
\subsection{\ourmethod makes GNN training way faster}\label{sec:exp_fast}
\begin{wrapfigure}{r}{0.5\textwidth}
\vspace{-1.3em}
  \begin{center}
    \includegraphics[width=0.5\textwidth]{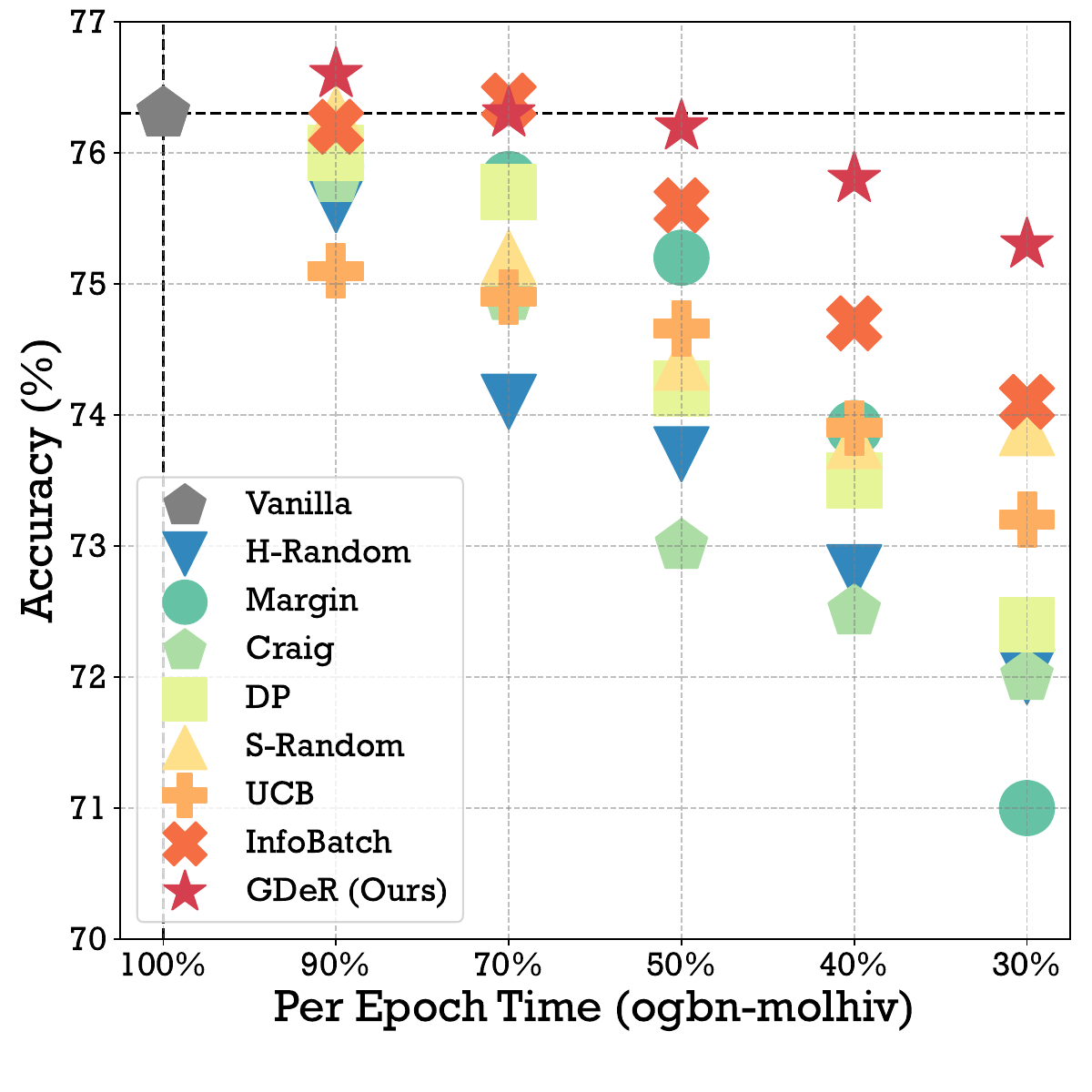}
  \end{center}
  \vspace{-2.3em}
  \caption{The trade-off between per epoch time and ROC-AUC (\%) of data pruning methods. Specifically, we report the test performance when pruning methods achieve per epoch times of $\{90\%, 70\%, 50\%, 40\%, 30\%\}$ of the full dataset training time. "Vanilla" denotes the original GNN backbone without any data pruning.}\label{fig:efficiency}
  \vspace{-2.3em}
\end{wrapfigure}

\vspace{-0.7em}
To answer \textbf{RQ1} and \textbf{RQ2}, we comprehensively compare \ourmethod with \textbf{fourteen} widely-used static pruning methods and \textbf{three} dynamic pruning methods, as outlined in \Cref{tab:rq1_ogb}, with more detailed explanations in \Cref{app:pruning_baseline}. Following~\cite {qin2023infobatch}, we add hard random and soft random pruning as baselines for a more comprehensive comparison. Specifically, we set the dataset remaining ratio $(1-s)\% \in \{20\%, 30\%, 50\%, 70\%\}$. The performance results are shown in \Cref{tab:rq1_ogb,tab:rq1_ogb_gps,tab:pretrain} and the efficiency comparisons are in \Cref{fig:efficiency}. Our observations (\textbf{Obs.}) are as summarized follows:

\newcommand{\blue}[1]{$_{\color{BlueGreen}\downarrow #1}$}
\newcommand{\red}[1]{$_{\color{RedOrange}\uparrow #1}$}
\begin{table*}[t]
    \caption{Performance comparison to state-of-the-art dataset pruning methods when remaining $\{20\%,30\%,50\%,70\%\}$ of the full set. All methods are trained using \textbf{PNA}, and the reported metrics represent the average of \textbf{five runs}. }
\label{tab:rq1_ogb}
    \centering
    \footnotesize
    \setlength{\tabcolsep}{3pt}
    % \begin{adjustbox}{max_width=\textwidth}
    \resizebox{\textwidth}{!}{
    \begin{tabular}{cc|cccc|cccc}
    \toprule
    \multirow{3}{*}{} & Dataset  & \multicolumn{4}{c|}{\textsc{ogbg-molhiv} (ROC-AUC $\uparrow$)} & \multicolumn{4}{c}{\textsc{ogbg-molpcba} (AP $\uparrow$)}  \\
    \midrule
    & Remaining Ratio \% & 20 & 30 & 50 & 70  & 20& 30 & 50 & 70 \\ \midrule
    \parbox[t]{4mm}{\multirow{14}{*}{\rotatebox[origin=c]{90}{Static}}}
    & Hard Random
    & 72.1\blue{4.2} & 72.4\blue{3.9} & 73.5\blue{2.8} & 75.6\blue{0.7} & 20.5\blue{7.6} & 22.9\blue{5.2} & 24.7\blue{3.4} & 28.0\blue{0.1}  \\
    & CD~\cite{agarwal2020contextual}
    & 71.9\blue{4.4} & 72.6\blue{3.7} & 73.8\blue{2.5} & 75.9\blue{0.4} 
    & 19.8\blue{8.3} & 22.6\blue{5.5} & 23.7\blue{4.4} & 27.8\blue{0.3} \\
    & Herding~\cite{welling2009herding}
    & 63.0\blue{13.3} & 64.9\blue{11.4} & 66.8\blue{9.5} & 75.2\blue{1.1}
    & 12.4\blue{15.7} & 14.0\blue{14.1} & 15.5\blue{12.6} & 21.8\blue{6.3}
    \\
    & K-Means~\cite{sener2018active}
    & 61.5\blue{14.8} & 65.9\blue{10.4} & 69.5\blue{6.8} & 74.7\blue{2.6} 
     & 18.5\blue{9.6} & 23.4\blue{4.7} & 23.2\blue{4.9} & 27.6\blue{0.5} \\
    & Least Confidence~\cite{coleman2019selection}
    & 72.1\blue{4.2} & 72.4\blue{3.9} & 75.6\blue{0.7} & 75.9\blue{0.4} & 21.0\blue{7.1} & 23.4\blue{4.7} & 25.0\blue{3.1} & 27.8\blue{0.3} \\
    & Margin~\cite{coleman2019selection}
    & 72.9\blue{3.4} & 71.3\blue{5.0} & 75.1\blue{1.2} & 76.0\blue{0.3}
    & 20.2\blue{7.9} & 23.3\blue{4.8} & 25.0\blue{3.1} & 28.3\red{0.2}
    \\
    & Forgetting~\cite{toneva2018empirical}
    & 72.6\blue{3.7} & 73.0\blue{3.3} & 73.9\blue{2.4} & 75.7\blue{0.6}
    & 20.7\blue{7.4} & 23.1\blue{5.0} & 24.1\blue{4.0} & 27.9\blue{0.2}
    \\
    & GraNd-4~\cite{paul2021deep}
    & 68.5\blue{7.8} & 72.7\blue{3.6} & 73.8\blue{2.5} & 75.7\blue{0.6}
    & 20.2\blue{7.9} & 22.9\blue{5.2} & 25.0\blue{3.1} & 28.0\blue{0.1}\\
    & GraNd-20~\cite{paul2021deep}
    & 74.7\blue{1.6} & 74.0\blue{2.3} & 74.9\blue{1.4} & 75.9\blue{0.4}  
    & 21.2\blue{6.9} & 23.8\blue{4.3} & 24.9\blue{3.2} & 27.8\blue{0.3} 
    \\
    & DeepFool~\cite{ducoffe2018adversarial}
    & 71.9\blue{4.4} & 72.5\blue{3.8} & 73.0\blue{3.3} & 75.6\blue{0.7} & 19.3\blue{8.8} & 22.7\blue{5.4} & 24.0\blue{4.1} & 27.7\blue{0.4} \\
    & Craig~\cite{mirzasoleiman2020coresets}
    & 71.8\blue{4.5} & 72.3\blue{4.0} & 73.5\blue{2.8} & 76.0\blue{0.3}  & 20.5\blue{7.6} & 23.1\blue{5.0} & 24.7\blue{3.4} & 27.8\blue{0.3}  \\
    & Glister~\cite{killamsetty2021glister}
    & 73.3\blue{3.0} & 74.4\blue{2.9} & 75.0\blue{1.3} & 76.2\blue{0.1}
    & 20.6\blue{7.5} & 23.4\blue{4.7} & 25.0\blue{3.1} & 27.9\blue{0.2}
    \\
    & Influence~\cite{koh2017understanding}
    & 71.5\blue{4.8} & 72.7\blue{3.6} & 73.5\blue{2.8} & 75.2\blue{1.1}
    & 19.7\blue{8.4} & 22.3\blue{5.8} & 23.9\blue{4.2} & 27.2\blue{0.9}\\
    & EL2N-2~\cite{toneva2018empirical}
    & 73.0\blue{3.3} & 74.5\blue{1.8} & 75.0\blue{1.3} & 76.1\blue{0.2} 
    & 20.9\blue{7.2} & 23.5\blue{4.6} & 24.3\blue{3.8} & 27.6\blue{0.5} \\
    % & EL2N-20~\cite{toneva2018empirical}
    % & 95.3\blue{0.3} & \textbf{95.1\blue{0.5}} & 91.9\blue{3.7} & 77.2\blue{1.0}& 72.1\blue{6.1} & - \\
    & DP~\cite{yang2023dataset}
    & 72.1\blue{4.2} & 73.5\blue{2.8} & 74.7\blue{1.6} & 76.0\blue{0.3}
    & 20.0\blue{8.1} & 22.7\blue{5.4} & 24.6\blue{3.5} & 27.7\blue{0.4}
    \\
    \midrule
    \parbox[t]{4mm}{\multirow{4}{*}{\rotatebox[origin=c]{90}{Dynamic}}}
    & Soft Random
    & 74.3\blue{2.0} & 73.9\blue{2.4} & 76.1\blue{0.2} & 76.2\blue{0.1} & 22.7\blue{5.4} & 24.8\blue{3.3} & 27.0\blue{1.1} & 27.8\blue{0.3} \\

    & $\epsilon$-greedy~\cite{raju2021ddp}
    & 73.8\blue{2.5} & 73.6\blue{2.7} & 75.6\blue{0.7} & 76.2\blue{0.1}
    & 24.0\blue{4.1} & 25.3\blue{2.8} & 27.1\blue{1.0} & 27.6\blue{0.5}
    \\

    & UCB~\cite{raju2021ddp}
    & 73.8\blue{2.5} & 73.7\blue{2.6} & 75.0\blue{1.3} & 75.8\blue{0.5}
    & 23.9\blue{4.2} & 25.8\blue{2.3} & 26.6\blue{1.5} & 28.1\red{0.0}
    \\

    & InfoBatch~\cite{qin2023infobatch}
    & 74.1\blue{2.2} & 74.0\blue{2.3} & 76.3\red{0.0} &  76.3\red{0.0} 
    & 24.1\blue{4.0} & 24.8\blue{3.3} & 27.3\blue{0.8} &  28.3\red{0.2} 
    \\

    & \ourmethod 
    & \colorbox[HTML]{DAE8FC}{\textbf{75.8\blue{\textbf{0.5}}}} 
    &  \colorbox[HTML]{DAE8FC}{\textbf{76.0\blue{\textbf{0.3}}}} 
    &   \colorbox[HTML]{DAE8FC}{\textbf{76.4\red{\textbf{0.1}}}} 
    & \colorbox[HTML]{DAE8FC}{\textbf{76.8\red{\textbf{0.5}}}}

    & \colorbox[HTML]{DAE8FC}{\textbf{24.8\blue{\textbf{3.3}}}} 
    &  \colorbox[HTML]{DAE8FC}{\textbf{26.0\blue{\textbf{2.1}}}} 
    &   \colorbox[HTML]{DAE8FC}{\textbf{28.0\blue{\textbf{0.1}}}} 
    & \colorbox[HTML]{DAE8FC}{\textbf{28.5\red{\textbf{0.4}}}}

    \\

    \midrule
    \multicolumn{2}{c|}{Whole Dataset} & \multicolumn{4}{c|}{76.3$_{\pm0.9}$} & \multicolumn{4}{c}{28.1$_{\pm0.3}$} \\
    \bottomrule
    \end{tabular}}
    \vspace{-1.5em}
    % \end{adjustbox}
\end{table*}

\vspace{-1em}
\paragraph{Obs.\ding{182} \ourmethod achieves maximum graph pruning with performance guarantees.} As shown in \Cref{tab:rq1_ogb,tab:rq1_ogb_gps}, \ourmethod consistently outperforms both static or dynamic baselines under various pruning ratios. On \textsc{ogbg-molhiv}+PNA, \ourmethod experiences only a $0.5\%$ performance decay even with $80\%$ pruning, surpassing the current state-of-the-art method InfoBatch, which suffers a $1.7\%$ decay. When pruning $50\%$ and $30\%$ of the data, \ourmethod even achieves performance improvements of $0.1\%$ and $0.5\%$, respectively.
\vspace{-1em}
\paragraph{Obs.\ding{183} The degree of redundancy varies across different datasets.} 
We observe that \textsc{ogbg-molpcba} is more sensitive to pruning than \textsc{ogbg-molhiv}, which suggests the degree of redundancy varies between datasets. For example, when pruning $80\%$ of the data, GraphGPS on \textsc{ogbg-molpcba} exhibits a performance decay ranging between $3.5\%\sim13.9\%$, significantly higher than the $2.5\%\sim11.5\%$ decay observed on \textsc{ogbg-molhiv}. However, as the remaining ratio increases, \ourmethod quickly recovers and surpasses the full dataset performance by $0.2\%$ at the $50\%$ pruning level.
\vspace{-1.2em}
\paragraph{Obs. \ding{184} \ourmethod can significantly accelerate GNN training.} \Cref{fig:efficiency} illustrates the per-epoch time and corresponding performance of each pruning method compared to full dataset training on \textsc{ogbg-molhiv}+GraphGPS. It is evident that \ourmethod can achieve a $2.0\times$ speedup without any performance loss (corresponding to $50\%$ per-epoch time). Even with a significant $3.3\times$ speedup, \ourmethod only experiences a moderate drop of $0.9\%$, which is superior to baselines including InfoBatch by a margin of $1.1\%\sim4.2\%$. Additionally, we observe from \Cref{tab:pretrain} that pretraining on ZINC with only $30\%$ of the data leads to a $1.53\%$ ROC-AUC improvement, with $2.81\times$ training time acceleration.

\begin{table*}[tp]
    \caption{Performance comparison to state-of-the-art dataset pruning methods. All methods are trained using \textbf{GraphGPS}, and the reported metrics represent the average of \textbf{five runs}. 
    }
\label{tab:rq1_ogb_gps}
    \centering
    \footnotesize
    \setlength{\tabcolsep}{3pt}
    % \begin{adjustbox}{max_width=\textwidth}
    \resizebox{\textwidth}{!}{
    \begin{tabular}{cc|cccc|cccc}
    \toprule
    \multirow{3}{*}{} & Dataset  & \multicolumn{4}{c|}{\textsc{ogbg-molhiv} (ROC-AUC $\uparrow$)} & \multicolumn{4}{c}{\textsc{ogbg-molpcba} (AP $\uparrow$)}  \\
    \midrule
    & Remaining Ratio \% & 20 & 30 & 50 & 70  & 20& 30 & 50 & 70 \\ \midrule
    \parbox[t]{4mm}{\multirow{17}{*}{\rotatebox[origin=c]{90}{Static}}}
    & Random
    & 69.3\blue{9.4} & 72.7\blue{6.0} & 73.4\blue{5.3} & 75.6\blue{3.1}  
    & 19.4\blue{7.8} & 21.7\blue{5.5} & 23.9\blue{3.3} & 26.3\blue{0.9}  \\
    
    & CD~\cite{agarwal2020contextual}
    & 72.6\blue{6.1} & 73.0\blue{5.7} & 75.3\blue{3.4} & 76.7\blue{2.0} 
    & 18.0\blue{9.2} & 20.7\blue{6.5} & 21.7\blue{5.5} & 26.4\blue{0.8} \\
    
    & Herding~\cite{welling2009herding}
    & 69.5\blue{9.2} & 73.3\blue{5.4} & 74.5\blue{4.2} & 75.9\blue{2.8}
    & 13.3\blue{13.9} & 14.0\blue{13.2} & 17.8\blue{9.4} & 23.0\blue{4.2} \\
    
    & K-Center~\cite{sener2018active}
    & 67.2\blue{11.5} & 70.8\blue{7.9} & 72.6\blue{6.1} & 73.9\blue{4.8} 
    & 16.9\blue{10.3} & 19.4\blue{7.8} & 22.8\blue{4.4} & 26.1\blue{1.1} \\
    
    & Least Confidence~\cite{coleman2019selection}
    & 73.9\blue{4.8} & 74.2\blue{4.5} & 75.8\blue{2.9} & 77.3\blue{1.4} 
    & 19.4\blue{7.6} & 21.9\blue{5.3} & 23.5\blue{3.7} & 26.0\blue{1.2} \\
    
    & Margin~\cite{coleman2019selection}
    & 74.0\blue{4.7} & 74.4\blue{4.3} & 75.8\blue{2.9} & 77.5\blue{1.2}
    & 18.8\blue{8.4} & 21.5\blue{5.7} & 23.9\blue{3.3} & 27.0\blue{0.2} \\
    
    & Forgetting~\cite{toneva2018empirical}
    & 74.2\blue{4.5} & 74.8\blue{3.9} & 75.6\blue{3.1} & 76.9\blue{1.8}
    & 18.3\blue{9.9} & 21.9\blue{5.3} & 23.3\blue{3.9} & 26.8\blue{0.4} \\
    
    & GraNd-4~\cite{paul2021deep}
    & 73.8\blue{4.9} & 74.2\blue{4.5} & 75.3\blue{3.4} & 77.5\blue{1.2} 
    & 18.0\blue{9.2} & 21.3\blue{5.9} & 23.6\blue{3.6} & 26.9\blue{0.3} \\
    
    & DeepFool~\cite{ducoffe2018adversarial}
    & 72.2\blue{6.5} & 73.3\blue{5.4} & 74.9\blue{3.8} & 75.5\blue{3.2} 
    & 17.6\blue{9.6} & 21.9\blue{5.3} & 23.2\blue{4.0} & 26.5\blue{0.7} \\
    
    & Craig~\cite{mirzasoleiman2020coresets}
    & 73.5\blue{5.2} & 74.4\blue{4.3} & 76.0\blue{2.7} & 77.9\blue{0.8}
    & 18.7\blue{8.5} & 22.7\blue{4.5} & 24.5\blue{2.7} & 27.1\blue{0.1}\\
    
    & Glister~\cite{killamsetty2021glister}
    & 73.6\blue{5.1} & 74.0\blue{4.7} & 75.8\blue{2.9} & 78.0\blue{0.7} 
    & 19.9\blue{7.3} & 22.5\blue{4.7} & 24.8\blue{2.4} & 27.0\red{0.2}\\
    
    & Influence~\cite{koh2017understanding}
    & 72.9\blue{5.8} & 73.7\blue{5.0} & 74.8\blue{3.9} & 77.4\blue{1.3} 
    & 17.7\blue{9.5} & 21.9\blue{5.3} & 23.5\blue{3.7} & 26.6\blue{0.6} \\
    
    % & EL2N-2~\cite{toneva2018empirical}
    % & 94.4\blue{1.2} & 93.2\blue{2.4} & 89.8\blue{5.8} & 74.1\blue{4.1} 
    % & 94.7\blue{0.9} & 94.1\blue{1.5} & 91.7\blue{3.9} & 75.3\blue{2.9} \\
    
    & EL2N-20~\cite{toneva2018empirical}
    & 74.0\blue{4.7} & {75.5\blue{3.2}} & 76.9\blue{1.8} & 77.7\blue{1.0}
    & 19.1\blue{8.1} & 22.9\blue{4.3} & 24.0\blue{3.2} & 26.0\blue{1.2} \\
    
    & DP~\cite{yang2023dataset}
    & 72.0\blue{6.7} & 74.1\blue{4.6} & 76.0\blue{2.7} & 76.9\blue{1.8} 
    & 19.6\blue{7.6} & 21.5\blue{5.7} & 24.9\blue{2.3} & 26.4\blue{0.8} \\
    
    \midrule
    \parbox[t]{4mm}{\multirow{4}{*}{\rotatebox[origin=c]{90}{Dynamic}}}
    & Soft Random
    & 74.0\blue{4.7} & 74.1\blue{4.6} & 74.4\blue{4.3} & 78.1\blue{0.6} 
    & 21.5\blue{5.7} & 22.4\blue{4.8} & 26.0\blue{1.2} & 27.1\blue{0.1} \\

    & $\epsilon$-greedy~\cite{raju2021ddp}
    & 74.7\blue{4.0} & 74.9\blue{3.8} & 76.6\blue{2.1} & 78.6\blue{0.1}
    & 22.8\blue{4.4} & 23.1\blue{4.1} & 26.3\blue{0.9} & 27.0\blue{0.2}
    \\

    & UCB~\cite{raju2021ddp}
    & 75.5\blue{3.2} & 74.9\blue{3.8} & 76.3\blue{2.4} & 78.0\blue{0.7}
    & 23.7\blue{3.5} & 24.1\blue{3.1} & 26.5\blue{0.7} & 27.2\red{0.0}\\

    & InfoBatch~\cite{qin2023infobatch}
    & 75.0\blue{3.4} & 75.6\blue{3.1} & 77.8\blue{0.9} & 78.5\blue{0.2}
    & 23.5\blue{3.7} & \colorbox[HTML]{DAE8FC}{\textbf{24.6\blue{2.6}}} & 26.7\blue{0.5} & 27.2\red{0.0} \\

    & \ourmethod 
    & \colorbox[HTML]{DAE8FC}{\textbf{76.5\blue{\textbf{2.2}}}} 
    & \colorbox[HTML]{DAE8FC}{\textbf{76.9\blue{\textbf{1.8}}}} 
    & \colorbox[HTML]{DAE8FC}{\textbf{78.7\red{\textbf{0.0}}}} 
    & \colorbox[HTML]{DAE8FC}{\textbf{79.1\red{\textbf{0.4}}}}
    & \colorbox[HTML]{DAE8FC}{\textbf{23.4\blue{\textbf{3.8}}}} 
    & {{24.5\blue{{2.7}}}} 
    & \colorbox[HTML]{DAE8FC}{\textbf{27.4\red{\textbf{0.2}}}} 
    & \colorbox[HTML]{DAE8FC}{\textbf{27.6\red{\textbf{0.4}}}} \\

    \midrule
    \multicolumn{2}{c|}{Whole Dataset} & \multicolumn{4}{c|}{78.7$_{\pm1.1}$} & \multicolumn{4}{c}{27.2$_{\pm0.2}$} \\
    \bottomrule
    \end{tabular}}
    % \end{adjustbox}
    \vspace{-1.2em}
\end{table*}

\subsection{\ourmethod Mitigates Graph Imbalance}\label{sec:exp_imbalance}
\vspace{-0.7em}
To answer \textbf{RQ3}, we tested \ourmethod in extremely imbalanced scenarios and compared its performance with other dynamic pruning methods. Following \cite{wang2022imbalanced}, we randomly set 25\%/25\% graphs as training/validation sets and within each of them, we designate one class as the minority class and reduce the number of graphs for this class in the training set (while increasing the others) until the imbalance ratio reached 1:9, which creates an extremely imbalanced scenario. The reported metrics are the average of 50 different data splits to avoid bias from data splitting. We observe from \Cref{fig:imbalance} that:

\begin{figure}[!htpb]
\setlength{\abovecaptionskip}{6pt}
\centering
\includegraphics[width=\textwidth]{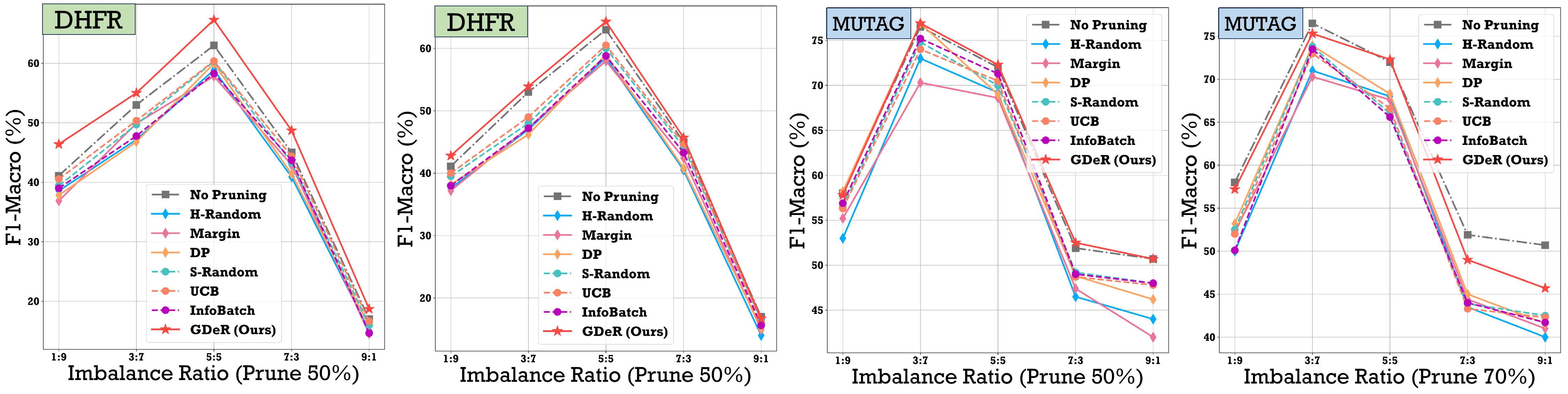}
\vspace{-1.8em}
\caption{Performance comparison of different pruning methods across various imbalance ratios. We utilize \textsc{mutag} and \textsc{dhfr} datasets with GCN, and reported the metrics when adjusting the imbalance ratios among \{1:9, 3:7, 5:5, 7:3, 9:1\}. ``No Pruning'' denotes training GCN without dataset pruning.} \label{fig:imbalance}
\vspace{-1.1em}
\end{figure}

\paragraph{Obs. \ding{185} \ourmethod can effectively mitigate imbalance issues.} As observed in \Cref{fig:imbalance}, baseline pruning methods struggle to outperform ``no-pruning'' GCN, resulting in substantial losses in speedup efficacy. In contrast, \ourmethod offers a more meaningful pruning approach. For instance, on \textsc{dhfr}, pruning 50\% of the data results in a $4.3\% $improvement in F1-Macro. This demonstrates that \ourmethod not only saves computational resources but also effectively mitigates data imbalance issues.

\vspace{-0.6em}
\subsection{\ourmethod Aids in GNN Robustness}
\label{sec:exp_robust}
\begin{figure}[!htpb]
\setlength{\abovecaptionskip}{6pt}
\centering
\includegraphics[width=\textwidth]{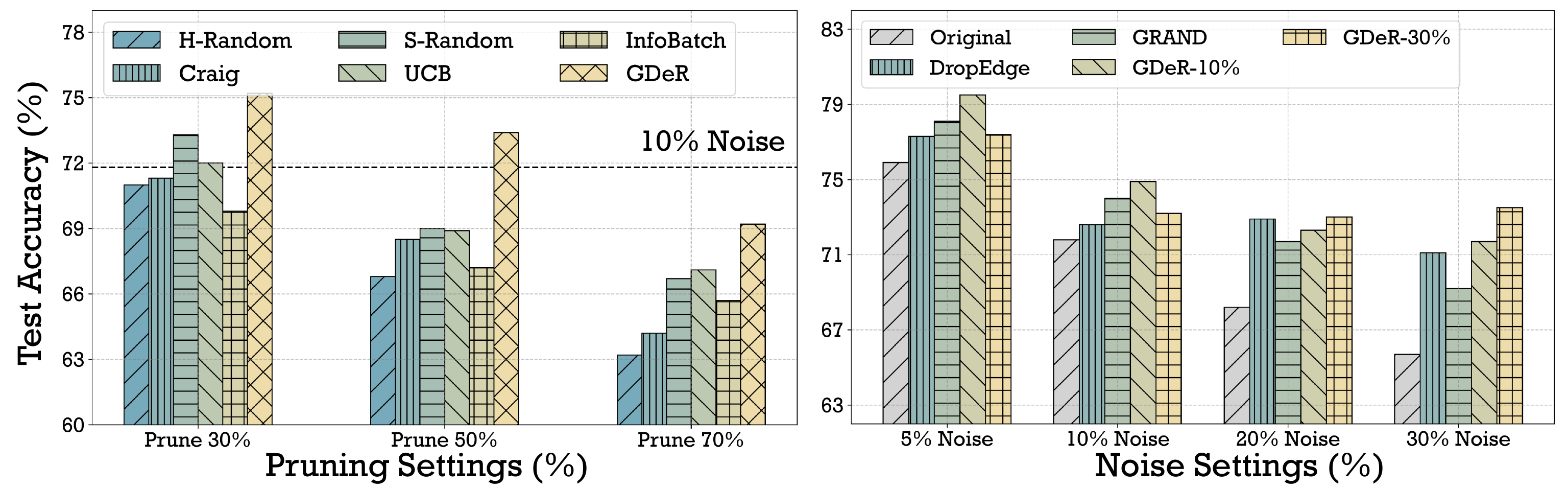}
\vspace{-1.8em}
\caption{(\textbf{\textit{Left}}) We report the performance of several top-performing pruning methods when perturbation noise is added to $10\%$ of the training set of \textsc{mutag}. The black dashed line represents the original GNN performance without pruning. (\textbf{\textit{Right}}) We compare \ourmethod with DropEdge and GRAND under different noise settings, utilizing \ourmethod with pruning ratios of $10\%$ and $30\%$.} \label{fig:noise}
\vspace{-1.em}
\end{figure}

% \vspace{-0.5em}
We divide \textbf{RQ4} into two sub-questions: (1) Is \ourmethod more robust to outlier perturbation compared to previous data pruning methods? (2) Can \ourmethod compete with mainstream methods designed to enhance GNN robustness? In practice, following \cite{li2022graphde}, we introduce perturbations to $k\%$ of the graph samples in the training set by adding Gaussian noise to the node features of the selected graphs. We compare \ourmethod against both data pruning baselines and GNN robustness enhancement baselines. The experimental results are presented in \Cref{fig:noise}, and we observe:
\vspace{-0.9em}
\paragraph{Obs. \ding{186} \ourmethod is a resource-saving GNN robustness booster.} From \Cref{fig:noise} (\textit{Left}), we observe that \ourmethod effectively counters noise perturbation, outperforming the GNN under outlier attacks at both $30\%$ and $50\%$ pruning rates. Notably, InfoBatch, which performed competitively in \textbf{RQ1}, suffers a significant performance drop ($2.0\%\sim6.1\%\downarrow$) in this biased training scenario, which is likely due to its loss magnitude-based sample selection mechanism, inadvertently amplifying the negative impact of high-loss outlier samples on the model. From \Cref{fig:noise} (\textit{Right}), we conclude that \ourmethod performs as well as or better than current robust GNN plugins, and it shows the most significant improvement in accuracy, with increases of $3.6\%$ and $7.8\%$ at noise ratios of $5\%$ and $30\%$, respectively.

\vspace{-0.9em}
\subsection{Ablation \& Sensitivity Study}\label{sec:ablation_sensi}
\vspace{-0.6em}
\paragraph{Ablation Study} To evaluate the effectiveness of the different modules in \ourmethod, we propose three variants: (1) \ourmethod w/o $\omega_e$, (2) \ourmethod w/o $\omega_r$, and (3) \ourmethod w/o $\omega_b$. \ourmethod w/o $\omega_e$ represents removing $\omega_e$ from \Cref{eq:final_distri}, with the other two variants defined similarly. We observe from \Cref{tab:ablation} that \ding{182} removing any component leads to a performance drop for \ourmethod, while removing $\omega_b$ in the imbalance scenario or $\omega_r$ in the biased scenario results in the most significant impact; \ding{183} \ourmethod w/o $\omega_e$ consistently underperforms across all scenarios, indicating that selecting highly representative samples is fundamental to the success of dynamic pruning methods.

\begin{minipage}[!t]{\linewidth}
    % \hspace{-0.5em}
    \begin{minipage}[!t]{0.48\linewidth}
    \centering
    \renewcommand{\arraystretch}{1.1}%  row spacing
    \scriptsize
    \tabcolsep=1.20mm
    \begin{tabular}{c|c|c|c}
    \toprule
    Setting & Normal & Imbalance  & Baised  \\
    \midrule
    \ourmethod & $84.21_{\pm 3.40}$  & ${76.32}_{\pm 4.70}$ & $77.84_{\pm 2.70}$  \\
    \ourmethod w/o $\omega_e$ & $79.77_{\pm 2.97}$  & ${73.78}_{\pm 2.96}$ & ${75.60}_{\pm 3.55}$ \\
    \ourmethod w/o $\omega_r$ & $84.01_{\pm 3.09}$  & $76.21_{\pm 3.42}$ & $77.96_{\pm 3.18}$  \\
    \ourmethod w/o $\omega_b$ & $83.46_{\pm 2.50}$  & $73.12_{\pm 2.50}$  & $75.02_{\pm 2.98}$ \\
    \bottomrule
    \end{tabular}
    \vspace{-0.5em}\makeatletter\def\@captype{table}\makeatother\caption{Ablation study on \ourmethod and its three variants. ``Imbalance'' refers to setting the imbalance ratio to be $\{1:9\}$, and ``Noisy'' refers to adding $5\%$ noise to the training set. All metrics are reported under $30\%$ pruning ratio.}\label{tab:ablation}%%
    \end{minipage}
    \hspace{0.1em}
    \begin{minipage}[!t]{0.47\linewidth}
    \centering
		\renewcommand{\arraystretch}{1.1}%  row spacing
    \scriptsize
    \tabcolsep=1.50mm
    \begin{tabular}{c|c|ccc}
    \toprule
    Ratio ($s\%$) & Metric  & $K=1$ & $K=2$ & $K=4$ \\
    \midrule
    \multirow{2}{*}{$20\%$} 
    & Perf. & $75.8_{\pm 1.5}$  & $\mathbf{76.5}_{\pm 1.4}$ & $76.1_{\pm 0.9}$\\
    & Time & $15.32$  & ${16.44}$ & $17.16$ \\
    
    \multirow{2}{*}{$50\%$} 
    & Perf. & $78.2_{\pm 1.4}$  & ${78.7}_{\pm 1.3}$ & $\mathbf{78.9}_{\pm 0.23}$\\
    & Time & $19.97$  & ${20.18}$ & $22.08$ \\
    
    \multirow{2}{*}{$70\%$} 
    & Perf. & $81.19_{\pm 2.0}$  & ${79.1}_{\pm 1.9}$ & $\mathbf{79.2}_{\pm 2.2}$\\
    & Time & $26.19$  & ${31.30}$ & $39.55$ \\
    \bottomrule
    \end{tabular}%
\vspace{-0.5em}    \makeatletter\def\@captype{table}\makeatother\caption{Sensitivity analysis on $K$. We report the ROCAUC (\%) and per-epoch time (s) on \textsc{ogbg-molhiv}+GraphGPS.}\label{tab:abla_k}%%
    \end{minipage}
\end{minipage}
\vspace{-1.1em}
\paragraph{Sensitivity and Efficiency Analysis} We investigate the impact of $K$, on the performance and efficiency of \ourmethod. Specifically, we vary $K \in \{1,2,4\}$ on \textsc{ogbg-molhiv}+GraphGPS and observe changes in performance and per-epoch time. We observe from \Cref{tab:abla_k} that $K=1$ leads to an under-learning of the hypersphere, resulting in consistently lower performance. While $K=4$ shows a marginal performance gain compared to $K=2$, for efficiency considerations, we opt for $K=2$ across all experiments. Additionally, we observe that data pruning significantly saves per-epoch time, with $s=20$ resulting in per-epoch times being $40\%\sim60\%$ of those achieved with $s=70$.

\vspace{-0.8em}
\section{Conclusion \& Future Work}
\vspace{-1em}
In this work, we propose the graph training debugging concept and explore soft dataset pruning in the graph learning area for the first time. Particularly, we present a prototype-guided soft pruning method, termed \ourmethod, which initially establishes a well-modeled graph embedding hypersphere and subsequently samples \textit{representative, balanced, and noise-free subsets} from this embedding space, debugging and troubleshooting graph processing. In the future, we plan to extend this concept to the CV realm, aiming to expedite the process of image training and provide efficient insights for the development of high-quality visual large-scale models.

\section*{Acknowledgement}
Dawei Cheng is supported by the National Natural Science Foundation of China (Grant No. 62102287). Yuxuan Liang is supported by the National Natural Science Foundation of China (No. 62402414).

\bibliography{ref}

\begin{thebibliography}{100}

\bibitem{motamedi2021data}
Mohammad Motamedi, Nikolay Sakharnykh, and Tim Kaldewey.
\newblock A data-centric approach for training deep neural networks with less data.
\newblock {\em arXiv preprint arXiv:2110.03613}, 2021.

\bibitem{zha2023data}
Daochen Zha, Zaid~Pervaiz Bhat, Kwei-Herng Lai, Fan Yang, Zhimeng Jiang, Shaochen Zhong, and Xia Hu.
\newblock Data-centric artificial intelligence: A survey.
\newblock {\em arXiv preprint arXiv:2303.10158}, 2023.

\bibitem{floridi2020gpt}
Luciano Floridi and Massimo Chiriatti.
\newblock Gpt-3: Its nature, scope, limits, and consequences.
\newblock {\em Minds and Machines}, 30:681--694, 2020.

\bibitem{li2024camel}
Guohao Li, Hasan Hammoud, Hani Itani, Dmitrii Khizbullin, and Bernard Ghanem.
\newblock Camel: Communicative agents for" mind" exploration of large language model society.
\newblock {\em Advances in Neural Information Processing Systems}, 36, 2024.

\bibitem{frantar2023sparsegpt}
Elias Frantar and Dan Alistarh.
\newblock Sparsegpt: Massive language models can be accurately pruned in one-shot.
\newblock In {\em International Conference on Machine Learning}, pages 10323--10337. PMLR, 2023.

\bibitem{ashkboos2024slicegpt}
Saleh Ashkboos, Maximilian~L Croci, Marcelo Gennari~do Nascimento, Torsten Hoefler, and James Hensman.
\newblock Slicegpt: Compress large language models by deleting rows and columns.
\newblock {\em arXiv preprint arXiv:2401.15024}, 2024.

\bibitem{sun2023simple}
Mingjie Sun, Zhuang Liu, Anna Bair, and J~Zico Kolter.
\newblock A simple and effective pruning approach for large language models.
\newblock {\em arXiv preprint arXiv:2306.11695}, 2023.

\bibitem{sanh2020movement}
Victor Sanh, Thomas Wolf, and Alexander Rush.
\newblock Movement pruning: Adaptive sparsity by fine-tuning.
\newblock {\em Advances in neural information processing systems}, 33:20378--20389, 2020.

\bibitem{touvron2023llama}
Hugo Touvron, Louis Martin, Kevin Stone, Peter Albert, Amjad Almahairi, Yasmine Babaei, Nikolay Bashlykov, Soumya Batra, Prajjwal Bhargava, Shruti Bhosale, et~al.
\newblock Llama 2: Open foundation and fine-tuned chat models.
\newblock {\em arXiv preprint arXiv:2307.09288}, 2023.

\bibitem{achiam2023gpt}
Josh Achiam, Steven Adler, Sandhini Agarwal, Lama Ahmad, Ilge Akkaya, Florencia~Leoni Aleman, Diogo Almeida, Janko Altenschmidt, Sam Altman, Shyamal Anadkat, et~al.
\newblock Gpt-4 technical report.
\newblock {\em arXiv preprint arXiv:2303.08774}, 2023.

\bibitem{wang2023chatvideo}
Junke Wang, Dongdong Chen, Chong Luo, Xiyang Dai, Lu~Yuan, Zuxuan Wu, and Yu-Gang Jiang.
\newblock Chatvideo: A tracklet-centric multimodal and versatile video understanding system, 2023.

\bibitem{chen2023minigptv2}
Jun Chen, Deyao Zhu, Xiaoqian Shen, Xiang Li, Zechu Liu, Pengchuan Zhang, Raghuraman Krishnamoorthi, Vikas Chandra, Yunyang Xiong, and Mohamed Elhoseiny.
\newblock Minigpt-v2: large language model as a unified interface for vision-language multi-task learning.
\newblock {\em arXiv preprint arXiv:2310.09478}, 2023.

\bibitem{zhu2023minigpt}
Deyao Zhu, Jun Chen, Xiaoqian Shen, Xiang Li, and Mohamed Elhoseiny.
\newblock Minigpt-4: Enhancing vision-language understanding with advanced large language models.
\newblock {\em arXiv preprint arXiv:2304.10592}, 2023.

\bibitem{chen2023vast}
Sihan Chen, Handong Li, Qunbo Wang, Zijia Zhao, Mingzhen Sun, Xinxin Zhu, and Jing Liu.
\newblock Vast: A vision-audio-subtitle-text omni-modality foundation model and dataset, 2023.

\bibitem{karlaš2022data}
Bojan Karlaš, David Dao, Matteo Interlandi, Bo~Li, Sebastian Schelter, Wentao Wu, and Ce~Zhang.
\newblock Data debugging with shapley importance over end-to-end machine learning pipelines, 2022.

\bibitem{shim2021core}
Jae-hun Shim, Kyeongbo Kong, and Suk-Ju Kang.
\newblock Core-set sampling for efficient neural architecture search.
\newblock {\em arXiv preprint arXiv:2107.06869}, 2021.

\bibitem{tukan2020coresets}
Murad Tukan, Alaa Maalouf, and Dan Feldman.
\newblock Coresets for near-convex functions.
\newblock {\em Advances in Neural Information Processing Systems}, 33:997--1009, 2020.

\bibitem{paul2021deep}
Mansheej Paul, Surya Ganguli, and Gintare~Karolina Dziugaite.
\newblock Deep learning on a data diet: Finding important examples early in training.
\newblock {\em Advances in Neural Information Processing Systems}, 34:20596--20607, 2021.

\bibitem{zhang2024graph}
Guibin Zhang, Kun Wang, Wei Huang, Yanwei Yue, Yang Wang, Roger Zimmermann, Aojun Zhou, Dawei Cheng, Jin Zeng, and Yuxuan Liang.
\newblock Graph lottery ticket automated.
\newblock In {\em The Twelfth International Conference on Learning Representations}, 2024.

\bibitem{wang2023brave}
Kun Wang, Yuxuan Liang, Xinglin Li, Guohao Li, Bernard Ghanem, Roger Zimmermann, Huahui Yi, Yudong Zhang, Yang Wang, et~al.
\newblock Brave the wind and the waves: Discovering robust and generalizable graph lottery tickets.
\newblock {\em IEEE Transactions on Pattern Analysis and Machine Intelligence}, 2023.

\bibitem{zhang2024heads}
Guibin Zhang, Yanwei Yue, Kun Wang, Junfeng Fang, Yongduo Sui, Kai Wang, Yuxuan Liang, Dawei Cheng, Shirui Pan, and Tianlong Chen.
\newblock Two heads are better than one: Boosting graph sparse training via semantic and topological awareness, 2024.

\bibitem{raju2021accelerating}
Ravi~S Raju, Kyle Daruwalla, and Mikko Lipasti.
\newblock Accelerating deep learning with dynamic data pruning.
\newblock {\em arXiv preprint arXiv:2111.12621}, 2021.

\bibitem{qin2023infobatch}
Ziheng Qin, Kai Wang, Zangwei Zheng, Jianyang Gu, Xiangyu Peng, Zhaopan Xu, Daquan Zhou, Lei Shang, Baigui Sun, Xuansong Xie, et~al.
\newblock Infobatch: Lossless training speed up by unbiased dynamic data pruning.
\newblock {\em arXiv preprint arXiv:2303.04947}, 2023.

\bibitem{zhao2023dataset}
Bo~Zhao and Hakan Bilen.
\newblock Dataset condensation with distribution matching.
\newblock In {\em Proceedings of the IEEE/CVF Winter Conference on Applications of Computer Vision}, pages 6514--6523, 2023.

\bibitem{wang2022cafe}
Kai Wang, Bo~Zhao, Xiangyu Peng, Zheng Zhu, Shuo Yang, Shuo Wang, Guan Huang, Hakan Bilen, Xinchao Wang, and Yang You.
\newblock Cafe: Learning to condense dataset by aligning features.
\newblock In {\em Proceedings of the IEEE/CVF Conference on Computer Vision and Pattern Recognition}, pages 12196--12205, 2022.

\bibitem{cazenavette2022dataset}
George Cazenavette, Tongzhou Wang, Antonio Torralba, Alexei~A Efros, and Jun-Yan Zhu.
\newblock Dataset distillation by matching training trajectories.
\newblock In {\em Proceedings of the IEEE/CVF Conference on Computer Vision and Pattern Recognition}, pages 4750--4759, 2022.

\bibitem{nguyen2021dataset}
Timothy Nguyen, Roman Novak, Lechao Xiao, and Jaehoon Lee.
\newblock Dataset distillation with infinitely wide convolutional networks.
\newblock {\em Advances in Neural Information Processing Systems}, 34:5186--5198, 2021.

\bibitem{li2023attend}
Xinglin Li, Kun Wang, Hanhui Deng, Yuxuan Liang, and Di~Wu.
\newblock Attend who is weak: Enhancing graph condensation via cross-free adversarial training.
\newblock {\em arXiv preprint arXiv:2311.15772}, 2023.

\bibitem{zhangnavigating}
Yuchen Zhang, Tianle Zhang, Kai Wang, Ziyao Guo, Yuxuan Liang, Xavier Bresson, Wei Jin, and Yang You.
\newblock Navigating complexity: Toward lossless graph condensation via expanding window matching.
\newblock In {\em Forty-first International Conference on Machine Learning}, 2024.

\bibitem{zhang2024two2}
Tianle Zhang, Yuchen Zhang, Kun Wang, Kai Wang, Beining Yang, Kaipeng Zhang, Wenqi Shao, Ping Liu, Joey~Tianyi Zhou, and Yang You.
\newblock Two trades is not baffled: Condense graph via crafting rational gradient matching.
\newblock {\em arXiv preprint arXiv:2402.04924}, 2024.

\bibitem{har2004coresets}
Sariel Har-Peled and Soham Mazumdar.
\newblock On coresets for k-means and k-median clustering.
\newblock In {\em Proceedings of the thirty-sixth annual ACM symposium on Theory of computing}, pages 291--300, 2004.

\bibitem{chen2009coresets}
Ke~Chen.
\newblock On coresets for k-median and k-means clustering in metric and euclidean spaces and their applications.
\newblock {\em SIAM Journal on Computing}, 39(3):923--947, 2009.

\bibitem{toneva2018empirical}
Mariya Toneva, Alessandro Sordoni, Remi Tachet~des Combes, Adam Trischler, Yoshua Bengio, and Geoffrey~J Gordon.
\newblock An empirical study of example forgetting during deep neural network learning.
\newblock {\em arXiv preprint arXiv:1812.05159}, 2018.

\bibitem{zhao2020dataset}
Bo~Zhao, Konda~Reddy Mopuri, and Hakan Bilen.
\newblock Dataset condensation with gradient matching.
\newblock {\em arXiv preprint arXiv:2006.05929}, 2020.

\bibitem{lu2023can}
Yao Lu, Xuguang Chen, Yuchen Zhang, Jianyang Gu, Tianle Zhang, Yifan Zhang, Xiaoniu Yang, Qi~Xuan, Kai Wang, and Yang You.
\newblock Can pre-trained models assist in dataset distillation?
\newblock {\em arXiv preprint arXiv:2310.03295}, 2023.

\bibitem{li2022graphde}
Zenan Li, Qitian Wu, Fan Nie, and Junchi Yan.
\newblock Graphde: A generative framework for debiased learning and out-of-distribution detection on graphs.
\newblock {\em Advances in Neural Information Processing Systems}, 35:30277--30290, 2022.

\bibitem{park2024robust}
Dongmin Park, Seola Choi, Doyoung Kim, Hwanjun Song, and Jae-Gil Lee.
\newblock Robust data pruning under label noise via maximizing re-labeling accuracy.
\newblock {\em Advances in Neural Information Processing Systems}, 36, 2024.

\bibitem{debnath1991structure}
Asim~Kumar Debnath, Rosa~L Lopez~de Compadre, Gargi Debnath, Alan~J Shusterman, and Corwin Hansch.
\newblock Structure-activity relationship of mutagenic aromatic and heteroaromatic nitro compounds. correlation with molecular orbital energies and hydrophobicity.
\newblock {\em J. Med. Chem.}, 34(2):786--797, 1991.

\bibitem{kipf2016}
Thomas~N. Kipf and Max Welling.
\newblock Semi-supervised classification with graph convolutional networks, 2016.

\bibitem{wang2022imbalanced}
Yu~Wang, Yuying Zhao, Neil Shah, and Tyler Derr.
\newblock Imbalanced graph classification via graph-of-graph neural networks.
\newblock In {\em Proceedings of the 31st ACM International Conference on Information \& Knowledge Management}, pages 2067--2076, 2022.

\bibitem{li2024rethinking}
Zhixun Li, Yushun Dong, Qiang Liu, and Jeffrey~Xu Yu.
\newblock Rethinking fair graph neural networks from re-balancing.
\newblock In {\em Proceedings of the 30th ACM SIGKDD Conference on Knowledge Discovery and Data Mining}, pages 1736--1745, 2024.

\bibitem{S3GCL_ICML24}
Guancheng Wan, Yijun Tian, Wenke Huang, Nitesh~V Chawla, and Mang Ye.
\newblock S3gcl: Spectral, swift, spatial graph contrastive learning.
\newblock In {\em Forty-first International Conference on Machine Learning}, 2024.

\bibitem{FGGP_AAAI24}
Guancheng Wan, Wenke Huang, and Mang Ye.
\newblock Federated graph learning under domain shift with generalizable prototypes.
\newblock In {\em Proceedings of the AAAI Conference on Artificial Intelligence}, volume~38, pages 15429--15437, 2024.

\bibitem{FedSSP_NeurIPS}
Zihan Tan, Guancheng Wan, Wenke Huang, and Mang Ye.
\newblock Fedssp: Federated graph learning with spectral knowledge and personalized preference.
\newblock In {\em Thirty-eighth Annual Conference on Neural Information Processing Systems}, 2024.

\bibitem{FGSSL_IJCAI23}
Wenke Huang, Guancheng Wan, Mang Ye, and Bo~Du.
\newblock Federated graph semantic and structural learning.
\newblock In {\em Proceedings of the Thirty-Second International Joint Conference on Artificial Intelligence}, pages 3830--3838, 2023.

\bibitem{arik2020protoattend}
Sercan~O Arik and Tomas Pfister.
\newblock Protoattend: Attention-based prototypical learning.
\newblock {\em Journal of Machine Learning Research}, 21(210):1--35, 2020.

\bibitem{li2020prototypical}
Junnan Li, Pan Zhou, Caiming Xiong, and Steven~CH Hoi.
\newblock Prototypical contrastive learning of unsupervised representations.
\newblock {\em arXiv preprint arXiv:2005.04966}, 2020.

\bibitem{hoffman2018cycada}
Judy Hoffman, Eric Tzeng, Taesung Park, Jun-Yan Zhu, Phillip Isola, Kate Saenko, Alexei Efros, and Trevor Darrell.
\newblock Cycada: Cycle-consistent adversarial domain adaptation.
\newblock In {\em International conference on machine learning}, pages 1989--1998. Pmlr, 2018.

\bibitem{tsai2018learning}
Yi-Hsuan Tsai, Wei-Chih Hung, Samuel Schulter, Kihyuk Sohn, Ming-Hsuan Yang, and Manmohan Chandraker.
\newblock Learning to adapt structured output space for semantic segmentation.
\newblock In {\em Proceedings of the IEEE conference on computer vision and pattern recognition}, pages 7472--7481, 2018.

\bibitem{pan2013graph}
Shirui Pan and Xingquan Zhu.
\newblock Graph classification with imbalanced class distributions and noise.
\newblock In {\em IJCAI}, pages 1586--1592, 2013.

\bibitem{hou2022graphmae}
Zhenyu Hou, Xiao Liu, Yukuo Cen, Yuxiao Dong, Hongxia Yang, Chunjie Wang, and Jie Tang.
\newblock Graphmae: Self-supervised masked graph autoencoders.
\newblock In {\em Proceedings of the 28th ACM SIGKDD Conference on Knowledge Discovery and Data Mining}, pages 594--604, 2022.

\bibitem{you2020graph}
Yuning You, Tianlong Chen, Yongduo Sui, Ting Chen, Zhangyang Wang, and Yang Shen.
\newblock Graph contrastive learning with augmentations.
\newblock {\em Advances in neural information processing systems}, 33:5812--5823, 2020.

\bibitem{you2021graph}
Yuning You, Tianlong Chen, Yang Shen, and Zhangyang Wang.
\newblock Graph contrastive learning automated.
\newblock In {\em International Conference on Machine Learning}, pages 12121--12132. PMLR, 2021.

\bibitem{wu2020comprehensive}
Zonghan Wu, Shirui Pan, Fengwen Chen, Guodong Long, Chengqi Zhang, and S~Yu Philip.
\newblock A comprehensive survey on graph neural networks.
\newblock {\em IEEE transactions on neural networks and learning systems}, 32(1):4--24, 2020.

\bibitem{zhou2020graph}
Jie Zhou, Ganqu Cui, Shengding Hu, Zhengyan Zhang, Cheng Yang, Zhiyuan Liu, Lifeng Wang, Changcheng Li, and Maosong Sun.
\newblock Graph neural networks: A review of methods and applications.
\newblock {\em AI open}, 1:57--81, 2020.

\bibitem{xiao2022graph}
Shunxin Xiao, Shiping Wang, Yuanfei Dai, and Wenzhong Guo.
\newblock Graph neural networks in node classification: survey and evaluation.
\newblock {\em Machine Vision and Applications}, 33(1):4, 2022.

\bibitem{zhang2018link}
Muhan Zhang and Yixin Chen.
\newblock Link prediction based on graph neural networks.
\newblock In {\em Proceedings of NIPS}, 2018.

\bibitem{liu2022graph}
Chuang Liu, Yibing Zhan, Jia Wu, Chang Li, Bo~Du, Wenbin Hu, Tongliang Liu, and Dacheng Tao.
\newblock Graph pooling for graph neural networks: Progress, challenges, and opportunities.
\newblock {\em arXiv preprint arXiv:2204.07321}, 2022.

\bibitem{wang2022searching}
Kun Wang, Yuxuan Liang, Pengkun Wang, Xu~Wang, Pengfei Gu, Junfeng Fang, and Yang Wang.
\newblock Searching lottery tickets in graph neural networks: A dual perspective.
\newblock In {\em The Eleventh International Conference on Learning Representations}, 2022.

\bibitem{chen2024uncovering}
Dingshuo Chen, Yanqiao Zhu, Jieyu Zhang, Yuanqi Du, Zhixun Li, Qiang Liu, Shu Wu, and Liang Wang.
\newblock Uncovering neural scaling laws in molecular representation learning.
\newblock {\em Advances in Neural Information Processing Systems}, 36, 2024.

\bibitem{koh2017understanding}
Pang~Wei Koh and Percy Liang.
\newblock Understanding black-box predictions via influence functions.
\newblock In {\em Proceedings of the 34th International Conference on Machine Learning-Volume 70}, pages 1885--1894. JMLR. org, 2017.

\bibitem{raju2021ddp}
Ravi~S Raju, Kyle Daruwalla, and Mikko Lipasti.
\newblock Accelerating deep learning with dynamic data pruning, 2021.

\bibitem{chen2024beyond}
Dingshuo Chen, Zhixun Li, Yuyan Ni, Guibin Zhang, Ding Wang, Qiang Liu, Shu Wu, Jeffrey~Xu Yu, and Liang Wang.
\newblock Beyond efficiency: Molecular data pruning for enhanced generalization.
\newblock {\em arXiv preprint arXiv:2409.01081}, 2024.

\bibitem{sui2021inductive}
Yongduo Sui, Xiang Wang, Tianlong Chen, Xiangnan He, and Tat-Seng Chua.
\newblock Inductive lottery ticket learning for graph neural networks.
\newblock 2021.

\bibitem{zeng2019graphsaint}
Hanqing Zeng, Hongkuan Zhou, Ajitesh Srivastava, Rajgopal Kannan, and Viktor Prasanna.
\newblock Graphsaint: Graph sampling based inductive learning method.
\newblock {\em arXiv preprint arXiv:1907.04931}, 2019.

\bibitem{zhang2023deep}
Yifan Zhang, Bingyi Kang, Bryan Hooi, Shuicheng Yan, and Jiashi Feng.
\newblock Deep long-tailed learning: A survey.
\newblock {\em IEEE Transactions on Pattern Analysis and Machine Intelligence}, 2023.

\bibitem{he2009learning}
Haibo He and Edwardo~A Garcia.
\newblock Learning from imbalanced data.
\newblock {\em IEEE Transactions on knowledge and data engineering}, 21(9):1263--1284, 2009.

\bibitem{chawla2002smote}
Nitesh~V Chawla, Kevin~W Bowyer, Lawrence~O Hall, and W~Philip Kegelmeyer.
\newblock Smote: synthetic minority over-sampling technique.
\newblock {\em Journal of artificial intelligence research}, 16:321--357, 2002.

\bibitem{kang2019decoupling}
Bingyi Kang, Saining Xie, Marcus Rohrbach, Zhicheng Yan, Albert Gordo, Jiashi Feng, and Yannis Kalantidis.
\newblock Decoupling representation and classifier for long-tailed recognition.
\newblock {\em arXiv preprint arXiv:1910.09217}, 2019.

\bibitem{lin2017focal}
Tsung-Yi Lin, Priya Goyal, Ross Girshick, Kaiming He, and Piotr Doll{\'a}r.
\newblock Focal loss for dense object detection.
\newblock In {\em Proceedings of the IEEE international conference on computer vision}, pages 2980--2988, 2017.

\bibitem{cui2019class}
Yin Cui, Menglin Jia, Tsung-Yi Lin, Yang Song, and Serge Belongie.
\newblock Class-balanced loss based on effective number of samples.
\newblock In {\em Proceedings of the IEEE/CVF conference on computer vision and pattern recognition}, pages 9268--9277, 2019.

\bibitem{tan2020equalization}
Jingru Tan, Changbao Wang, Buyu Li, Quanquan Li, Wanli Ouyang, Changqing Yin, and Junjie Yan.
\newblock Equalization loss for long-tailed object recognition.
\newblock In {\em Proceedings of the IEEE/CVF conference on computer vision and pattern recognition}, pages 11662--11671, 2020.

\bibitem{menon2020long}
Aditya~Krishna Menon, Sadeep Jayasumana, Ankit~Singh Rawat, Himanshu Jain, Andreas Veit, and Sanjiv Kumar.
\newblock Long-tail learning via logit adjustment.
\newblock {\em arXiv preprint arXiv:2007.07314}, 2020.

\bibitem{DRGCN}
Min Shi, Yufei Tang, Xingquan Zhu, David Wilson, and Jianxun Liu.
\newblock Multi-class imbalanced graph convolutional network learning.
\newblock In {\em IJCAI}, 2020.

\bibitem{graphsmote}
Tianxiang Zhao, Xiang Zhang, and Suhang Wang.
\newblock Graphsmote: Imbalanced node classification on graphs with graph neural networks.
\newblock In {\em WSDM}, 2021.

\bibitem{qu2021imgagn}
Liang Qu, Huaisheng Zhu, Ruiqi Zheng, Yuhui Shi, and Hongzhi Yin.
\newblock Imgagn: Imbalanced network embedding via generative adversarial graph networks.
\newblock In {\em Proceedings of the 27th ACM SIGKDD Conference on Knowledge Discovery \& Data Mining}, pages 1390--1398, 2021.

\bibitem{ma2020towards}
Jiaqi Ma, Shuangrui Ding, and Qiaozhu Mei.
\newblock Towards more practical adversarial attacks on graph neural networks.
\newblock {\em Advances in neural information processing systems}, 33:4756--4766, 2020.

\bibitem{zhu2019robust}
Dingyuan Zhu, Ziwei Zhang, Peng Cui, and Wenwu Zhu.
\newblock Robust graph convolutional networks against adversarial attacks.
\newblock In {\em Proceedings of the 25th ACM SIGKDD international conference on knowledge discovery \& data mining}, pages 1399--1407, 2019.

\bibitem{li2024gslb}
Zhixun Li, Xin Sun, Yifan Luo, Yanqiao Zhu, Dingshuo Chen, Yingtao Luo, Xiangxin Zhou, Qiang Liu, Shu Wu, Liang Wang, et~al.
\newblock Gslb: The graph structure learning benchmark.
\newblock {\em Advances in Neural Information Processing Systems}, 36, 2024.

\bibitem{chen2017practical}
Yizheng Chen, Yacin Nadji, Athanasios Kountouras, Fabian Monrose, Roberto Perdisci, Manos Antonakakis, and Nikolaos Vasiloglou.
\newblock Practical attacks against graph-based clustering.
\newblock In {\em Proceedings of the 2017 ACM SIGSAC conference on computer and communications security}, pages 1125--1142, 2017.

\bibitem{wang2023turning}
Binghui Wang, Meng Pang, and Yun Dong.
\newblock Turning strengths into weaknesses: A certified robustness inspired attack framework against graph neural networks.
\newblock In {\em Proceedings of the IEEE/CVF Conference on Computer Vision and Pattern Recognition}, pages 16394--16403, 2023.

\bibitem{yang2023graphguard}
Han Yang, Binghui Wang, Jinyuan Jia, et~al.
\newblock Graphguard: Provably robust graph classification against adversarial attacks.
\newblock In {\em The Twelfth International Conference on Learning Representations}, 2023.

\bibitem{zhang2021backdoor}
Zaixi Zhang, Jinyuan Jia, Binghui Wang, and Neil~Zhenqiang Gong.
\newblock Backdoor attacks to graph neural networks.
\newblock In {\em Proceedings of the 26th ACM Symposium on Access Control Models and Technologies}, pages 15--26, 2021.

\bibitem{wang2020understanding}
Tongzhou Wang and Phillip Isola.
\newblock Understanding contrastive representation learning through alignment and uniformity on the hypersphere.
\newblock In {\em Proceedings of ICML}, 2020.

\bibitem{khosla2020supervised}
Prannay Khosla, Piotr Teterwak, Chen Wang, Aaron Sarna, Yonglong Tian, Phillip Isola, Aaron Maschinot, Ce~Liu, and Dilip Krishnan.
\newblock Supervised contrastive learning.
\newblock {\em Advances in neural information processing systems}, 33:18661--18673, 2020.

\bibitem{lu2024learning}
Haodong Lu, Dong Gong, Shuo Wang, Jason Xue, Lina Yao, and Kristen Moore.
\newblock Learning with mixture of prototypes for out-of-distribution detection.
\newblock {\em arXiv preprint arXiv:2402.02653}, 2024.

\bibitem{du2022siren}
Xuefeng Du, Gabriel Gozum, Yifei Ming, and Yixuan Li.
\newblock Siren: Shaping representations for detecting out-of-distribution objects.
\newblock {\em Advances in Neural Information Processing Systems}, 35:20434--20449, 2022.

\bibitem{sehwag2021ssd}
Vikash Sehwag, Mung Chiang, and Prateek Mittal.
\newblock Ssd: A unified framework for self-supervised outlier detection.
\newblock {\em arXiv preprint arXiv:2103.12051}, 2021.

\bibitem{abbas2024effective}
Amro Abbas, Evgenia Rusak, Kushal Tirumala, Wieland Brendel, Kamalika Chaudhuri, and Ari~S Morcos.
\newblock Effective pruning of web-scale datasets based on complexity of concept clusters.
\newblock {\em arXiv preprint arXiv:2401.04578}, 2024.

\bibitem{frigui2004unsupervised}
Hichem Frigui and Olfa Nasraoui.
\newblock Unsupervised learning of prototypes and attribute weights.
\newblock {\em Pattern recognition}, 37(3):567--581, 2004.

\bibitem{sutherland2003spline}
Jeffrey~J Sutherland, Lee~A O'brien, and Donald~F Weaver.
\newblock Spline-fitting with a genetic algorithm: A method for developing classification structure- activity relationships.
\newblock {\em J Chem Inform Comput Sci}, 43(6), 2003.

\bibitem{hu2020open}
Weihua Hu, Matthias Fey, Marinka Zitnik, Yuxiao Dong, Hongyu Ren, Bowen Liu, Michele Catasta, and Jure Leskovec.
\newblock Open graph benchmark: Datasets for machine learning on graphs.
\newblock {\em arXiv preprint arXiv:2005.00687}, 2020.

\bibitem{irwin2012zinc}
John~J Irwin, Teague Sterling, Michael~M Mysinger, Erin~S Bolstad, and Ryan~G Coleman.
\newblock {ZINC}: A free tool to discover chemistry for biology.
\newblock {\em Journal of Chemical Information and Modeling}, 52(7):1757--1768, 2012.

\bibitem{rampavsek2022recipe}
Ladislav Ramp{\'a}{\v{s}}ek, Michael Galkin, Vijay~Prakash Dwivedi, Anh~Tuan Luu, Guy Wolf, and Dominique Beaini.
\newblock Recipe for a general, powerful, scalable graph transformer.
\newblock {\em Advances in Neural Information Processing Systems}, 35:14501--14515, 2022.

\bibitem{kipf2016semi}
Thomas~N. Kipf and Max Welling.
\newblock {Semi-Supervised Classification with Graph Convolutional Networks}.
\newblock In {\em Proceedings of the 5th International Conference on Learning Representations}, 2017.

\bibitem{corso2020principal}
Gabriele Corso, Luca Cavalleri, Dominique Beaini, Pietro Li{\`o}, and Petar Veli{\v{c}}kovi{\'c}.
\newblock Principal neighbourhood aggregation for graph nets.
\newblock {\em Advances in Neural Information Processing Systems}, 33:13260--13271, 2020.

\bibitem{agarwal2020contextual}
Sharat Agarwal, Himanshu Arora, Saket Anand, and Chetan Arora.
\newblock Contextual diversity for active learning.
\newblock In {\em ECCV}, pages 137--153. Springer, 2020.

\bibitem{welling2009herding}
Max Welling.
\newblock Herding dynamical weights to learn.
\newblock In {\em ICMLg}, pages 1121--1128, 2009.

\bibitem{sener2018active}
Ozan Sener and Silvio Savarese.
\newblock Active learning for convolutional neural networks: A core-set approach.
\newblock In {\em ICLR}, 2018.

\bibitem{coleman2019selection}
Cody Coleman, Christopher Yeh, Stephen Mussmann, Baharan Mirzasoleiman, Peter Bailis, Percy Liang, Jure Leskovec, and Matei Zaharia.
\newblock Selection via proxy: Efficient data selection for deep learning.
\newblock In {\em ICLR}, 2019.

\bibitem{ducoffe2018adversarial}
Melanie Ducoffe and Frederic Precioso.
\newblock Adversarial active learning for deep networks: a margin based approach.
\newblock {\em arXiv preprint arXiv:1802.09841}, 2018.

\bibitem{mirzasoleiman2020coresets}
Baharan Mirzasoleiman, Jeff Bilmes, and Jure Leskovec.
\newblock Coresets for data-efficient training of machine learning models.
\newblock In {\em ICML}. PMLR, 2020.

\bibitem{killamsetty2021glister}
Krishnateja Killamsetty, Durga Sivasubramanian, Ganesh Ramakrishnan, and Rishabh Iyer.
\newblock Glister: Generalization based data subset selection for efficient and robust learning.
\newblock In {\em Proceedings of the AAAI Conference on Artificial Intelligence}, 2021.

\bibitem{yang2023dataset}
Shuo Yang, Zeke Xie, Hanyu Peng, Min Xu, Mingming Sun, and Ping Li.
\newblock Dataset pruning: Reducing training data by examining generalization influence.
\newblock In {\em The Eleventh International Conference on Learning Representations}, 2023.

\bibitem{choromanski2020rethinking}
Krzysztof Choromanski, Valerii Likhosherstov, David Dohan, Xingyou Song, Andreea Gane, Tamas Sarlos, Peter Hawkins, Jared Davis, Afroz Mohiuddin, Lukasz Kaiser, et~al.
\newblock Rethinking attention with performers.
\newblock {\em arXiv preprint arXiv:2009.14794}, 2020.

\bibitem{hu2019strategies}
Weihua Hu, Bowen Liu, Joseph Gomes, Marinka Zitnik, Percy Liang, Vijay Pande, and Jure Leskovec.
\newblock Strategies for pre-training graph neural networks.
\newblock {\em arXiv preprint arXiv:1905.12265}, 2019.

\bibitem{zhu2017prune}
Michael Zhu and Suyog Gupta.
\newblock To prune, or not to prune: exploring the efficacy of pruning for model compression.
\newblock {\em arXiv preprint arXiv:1710.01878}, 2017.

\bibitem{sterling2015zinc}
Teague Sterling and John~J Irwin.
\newblock Zinc 15--ligand discovery for everyone.
\newblock {\em Journal of Chemical Information and Modeling}, 55(11):2324--2337, 2015.

\bibitem{wu2018moleculenet}
Zhenqin Wu, Bharath Ramsundar, Evan~N Feinberg, Joseph Gomes, Caleb Geniesse, Aneesh~S Pappu, Karl Leswing, and Vijay Pande.
\newblock {MoleculeNet}: A benchmark for molecular machine learning.
\newblock {\em Chemical Science}, 9(2):513--530, 2018.

\bibitem{wu2018unsupervised}
Zhirong Wu, Yuanjun Xiong, Stella~X Yu, and Dahua Lin.
\newblock Unsupervised feature learning via non-parametric instance discrimination.
\newblock In {\em Proceedings of the IEEE conference on computer vision and pattern recognition}, pages 3733--3742, 2018.

\bibitem{oord2018representation}
Aaron van~den Oord, Yazhe Li, and Oriol Vinyals.
\newblock Representation learning with contrastive predictive coding.
\newblock {\em arXiv preprint arXiv:1807.03748}, 2018.

\bibitem{he2020momentum}
Kaiming He, Haoqi Fan, Yuxin Wu, Saining Xie, and Ross Girshick.
\newblock Momentum contrast for unsupervised visual representation learning.
\newblock In {\em Proceedings of CVPR}, 2020.

\bibitem{chen2020simple}
Ming Chen, Zhewei Wei, Zengfeng Huang, Bolin Ding, and Yaliang Li.
\newblock Simple and deep graph convolutional networks.
\newblock In {\em International conference on machine learning}, pages 1725--1735. PMLR, 2020.

\bibitem{robinson2021contrastive}
Joshua~David Robinson, Ching-Yao Chuang, Suvrit Sra, and Stefanie Jegelka.
\newblock Contrastive learning with hard negative samples.
\newblock In {\em International Conference on Learning Representations}, 2021.

\bibitem{assran2021semi}
Mahmoud Assran, Mathilde Caron, Ishan Misra, Piotr Bojanowski, Armand Joulin, Nicolas Ballas, and Michael Rabbat.
\newblock Semi-supervised learning of visual features by non-parametrically predicting view assignments with support samples.
\newblock In {\em Proceedings of the IEEE/CVF International Conference on Computer Vision}, pages 8443--8452, 2021.

\bibitem{wang2021understanding}
Feng Wang and Huaping Liu.
\newblock Understanding the behaviour of contrastive loss.
\newblock In {\em Proceedings of the IEEE/CVF conference on computer vision and pattern recognition}, pages 2495--2504, 2021.

\bibitem{caron2018deep}
Mathilde Caron, Piotr Bojanowski, Armand Joulin, and Matthijs Douze.
\newblock Deep clustering for unsupervised learning of visual features.
\newblock In {\em Proceedings of the European conference on computer vision (ECCV)}, pages 132--149, 2018.

\bibitem{snell2017prototypical}
Jake Snell, Kevin Swersky, and Richard Zemel.
\newblock Prototypical networks for few-shot learning.
\newblock {\em Advances in neural information processing systems}, 30, 2017.

\bibitem{li2021prototypical}
Junnan Li, Pan Zhou, Caiming Xiong, and Steven Hoi.
\newblock Prototypical contrastive learning of unsupervised representations.
\newblock In {\em International Conference on Learning Representations}, 2021.

\end{thebibliography}
\bibliographystyle{unsrt}

% \section{Appendix / supplemental material}
\begin{appendices}
\crefalias{section}{appendix}
\crefalias{subsection}{appendix}

\section{Notations}
We conclude the commonly used notations throughout the manuscript in \Cref{tab:notation}.
\begin{table}[!htbp]\footnotesize
  \centering
  \caption{The notations that are commonly used in the manuscript.}
   \setlength{\tabcolsep}{3pt} %biao kuan 
   \renewcommand\arraystretch{1.3} %
  \vspace{1em}
    \begin{tabular}{cc}
    \toprule
    Notation & Definition \\
    \midrule
    $\mathcal{G} = \left\{\mathcal{V}, \mathcal{E} \right\} = \left\{  \mathbf{A},\mathbf{X}\right\}$          & Input graph \\
    $\mathbf{A}$  & Input adjacency matrix \\
    $\mathbf{X}$  & Node features \\
    $\mathcal{D} = \{z_i\}_{i=1}^{|\mathcal{D}|} = \{(\mathcal{G}_i, \mathbf{Y}_i)\}_{i=1}^{|\mathcal{D}|}$ & Graph datasets \\
    $\mathbf{h}_i$ & Graph embedding for $\mathcal{G}_i$\\
    $f_\theta$ & GNN encoder \\
    $g_\phi$ & Feature projector \\
    $\mathbf{P}^c = \{\mathbf{p}_k^c\}_{k=1}^K$ & Total $K$ Prototypes for class $c$\\
    $\mathcal{X}_t$ & Remained training set at the $t$-th epoch \\
    $\Tilde{\mathcal{X}}_t$ & Pruned training set at the $t$-th epoch \\
    $\mathbf{z}_i \in \mathbb{R}^D$ & Projected embedding for $\mathcal{G}_i$ \\
    $\omega_{\text{r}}(\mathbf{z}_i)$ & Outlier risk assessment metric \\
    $\omega_{\text{e}}(\mathbf{z}_i)$ & Sample familiarity matric \\
    $\omega_{\text{b}}(\mathbf{z}_i)$ & Sample balancing score \\
    $\omega(\mathbf{z}_i)$ & Sampling probability for $\mathbf{z}_i$ \\
    ${\Psi}(t)$ & Scheduler function that controls how many samples to choose from $\mathcal{X}_t$ \\
    $\Tilde{\Psi}(t)$ & Scheduler function that controls how many samples to choose from $\mathcal{X}_{t+1}$ \\

    \bottomrule
    \end{tabular}%
  \label{tab:notation}%
\end{table}%

\section{Experimental Details}
\subsection{Dataset Details}\label{app:exp_dataset}

The graph dataset details are summarized in \Cref{tab:dataset}.

\begin{table}[htb]
\centering
% \vspace{-4mm}
\caption{Graph datasets statistics.}
% \vspace{-0.5em}
\label{tab:dataset}
% \resizebox{\columnwidth}{!}{
\begin{tabular}{c  |c |c |c  |c | c  }
\toprule
Dataset  & $\#$Graph &  $\#$Node & $\#$Edge & $\#$Classes & Metric  \\ 
\midrule
\textsc{mutag} & 188 & 17.93 & 19.79   & 2 & {Accuracy}/F1-macro \\
\textsc{dhfr} & 756 & 42.43	 & 44.54   & 2 & {Accuracy}/F1-macro \\
\midrule
\textsc{ogbg-molhiv} & 41,127	 & 25.5 & 27.5   & 40 & ROC-AUC \\
\textsc{ogbg-molpcba} & 	437,929	 & 26.0 & 28.1   & 2 & Average Precision \\ \midrule
\textsc{zinc15} & 70,100 & 50.5	 &  564.5& {10} & {Accuracy} \\
\bottomrule
\end{tabular}
%}
\vspace{-2mm}
\end{table}

\subsection{Backbone Settings}\label{app:exp_backbone}

We choose three representative GNNs, including one classic message-passing network GCN~\cite{kipf2016semi}, one classical graph classification backbone PNA~\cite{corso2020principal} and a graph transformer backbone GraphGPS~\cite{rampavsek2022recipe}. 
For GCN, we simply set $layer\_num=3$ and $\texttt{hidden\_dim}=128$. For PNA, we set $layer\_num=4$, and $\texttt{hidden\_dim}=64$, $\texttt{edge\_dim}=16$. The rest configurations are the same as provided by \cite{corso2020principal} (\url{https://github.com/lukecavabarrett/pna/blob/master/models/pytorch_geometric/example.py}). For GraphGPS, we uniformly set $\texttt{hidden\_dim}=64, \texttt{pe\_dim}=8$ and utilize random walk encoding, {performer}~\cite{choromanski2020rethinking}, and GINE~\cite{hu2019strategies} as the positional encoding, attention module and the local convolutional module, respectively. The rest configurations are the same as provided by the PyTorch library (\url{https://github.com/pyg-team/pytorch_geometric/blob/master/examples/graph_gps.py}). All the experiments are conducted on NVIDIA Tesla V100 (32GB GPU), using PyTorch and PyTorch Geometric framework.

\subsection{Pruning Baselines}\label{app:pruning_baseline}

As for static pruning methods, we first introduce Hard Random, which conducts a random sample selection before training. Influence~\cite{koh2017understanding} and EL2N~\cite{paul2021deep} are two classical static pruning methods that prune samples based on Influence-score and EL2N-score, respectively. DP~\cite{yang2023dataset} conducts pruning with consideration of generalization. Following the methodology of ~\cite{qin2023infobatch}, we introduce a total of 13 static data pruning methods. These methods select a core set of data via predefined score functions or heuristic knowledge. Additionally, we introduce three dynamic pruning methods, including $\epsilon$-greedy~\cite{raju2021ddp}, UCB~\cite{raju2021ddp}, and InfoBatch~\cite{qin2023infobatch}. Following ~\cite{raju2021ddp,qin2023infobatch}, we also introduce the dynamic pruning baseline, termed Soft Random, which conducts random selection in each epoch.

\subsection{Metrics}

For \textsc{mutag} and \textsc{dhfr}, the metrics used vary across different scenarios. In the normal (\Cref{sec:exp_fast}) and biased (\Cref{sec:exp_robust}) scenarios, we use accuracy. However, in the imbalanced scenario (\Cref{sec:exp_imbalance}), accuracy does not faithfully reflect the performance for the minority group. Following previous works in imbalanced classification~\cite{graphsmote}, we choose to use F1-macro, which computes the accuracy independently for each class and then takes the average, treating different classes equally. For \textsc{ogbg-molhiv} and \textsc{ogbg-molpcba}, we use ROC-AUC and Average Precision (AP), following~\cite{hu2020open}.

\subsection{Scheduler Function}\label{app:scheduler}
${\Psi}(t)$ and $\Tilde{\Psi}(t)$ are scheduler functions that determine the proportions of samples in $\mathcal{X}_{t+1}$ originating from $\mathcal{X}_t$ and $\Tilde{\mathcal{X}}_t$, respectively. For simplicity, we adopt the Inverse Power function~\cite{zhu2017prune}:
\begin{equation}\label{eq:scheduler}
{\Psi}(t) = |\mathcal{X}_t| \cdot\varsigma \left(1 - \frac{t}{T}\right)^\kappa,\; \Tilde{\Psi}(t) = |\mathcal{X}_t| - {\Psi}(t) 
\end{equation}
where $\varsigma$ denotes the initial ratio and $\kappa$ is the decay factor controlling the rate at which the ratio decreases over intervals. In practice, we uniformly set $\varsigma=0.7$ and $\kappa=2$.

\section{Algorithm Workflow}\label{app:algo}

The algorithm framework is presented in Algo.~\ref{alg:algo}.
\begin{algorithm}[!htpb]
\caption{Algorithm workflow of \ourmethod}\label{alg:algo}
\Input{Graph datasets $\mathcal{D} = \{z_i\}_{i=1}^{|\mathcal{D}|} = \{(\mathcal{G}_i, \mathbf{Y}_i)\}_{i=1}^{|\mathcal{D}|}$, the number of epochs $T$, GNN encoder $f_\theta$, feature projector $g_\phi$, }

Initialized $M$ prototypes $\{\mathbf{p}^{(C)}_k\}_{i=1}^M$ for class C

\For{\rm{epoch} $t \leftarrow 1$ \KwTo $T$}{

\tcc{\textcolor{blue}{Extract graph-level embedding and Projection}}

$\mathcal{X}_t \leftarrow$ current training set.

\For{\rm{sample index} $i \leftarrow 1$ \KwTo $|\mathcal{X}_t|$}{
Compute graph embedding $\mathbf{h}_i \leftarrow g_\theta(\mathcal{G}_i)$.

Project graph embedding onto hypersphere by $\mathbf{z}_i = \mathbf{z}'/||\mathbf{z}'||_2, \mathbf{z}' = g_\theta(\mathbf{h}_i)$.
}
% % \tcc{\textcolor{blue}{Ego-graph decomposition}}
% Decompose $\mathcal{G}$ into ego-graph representations $\{\mathcal{G}^{(1)}, \mathcal{G}^{(2)}, \cdots, \mathcal{G}^{(N)}\}$.

Determine which prototype cluster $\chi_c$ each graph sample $\mathbf{z}_i$ corresponds to\Comment*[r]{\textcolor{blue}{Eq.~\ref{eq:mixture_vmf}}}

\tcc{\textcolor{blue}{Formatting sampling distribution}}
Calculate the outlier score $\omega_{\text{r}}(\mathbf{z}_i)$ by prototype-based Mahalanobis distance\Comment*[r]{\textcolor{blue}{Eq.~\ref{eq:outlier}}}

Calculate the familiarity score $\omega_{\text{e}}(\mathbf{z}_i)$ based on prorotype-sample distance\Comment*[r]{\textcolor{blue}{Eq.~\ref{eq:familarity}}}

Calculate the balancing distribution  $\omega_{\text{b}}$ based on cluster volume\Comment*[r]{\textcolor{blue}{Eq.~\ref{eq:balance}}}

Formulate sampling distribution $ \omega(\mathbf{z}_i) = \frac{{\omega_{\text{e}}^\sigma(\mathbf{z}_i)}}{ \left(\omega_{\text{r}}^{\sigma}(\mathbf{z}_i)+\epsilon\right)\cdot \left(\omega_{\text{b}}^\sigma(\mathbf{z}_i)+\epsilon\right)}$\Comment*[r]{\textcolor{blue}{Eq.~\ref{eq:final_distri}}}

\tcc{\textcolor{blue}{Dataset sampling}}

Initialize the sample set for the $(t+1)$-th epoch $\mathcal{X}_{t+1}\leftarrow \emptyset$

\tcc{\textcolor{blue}{Sample from currently remained set}}
$\mathcal{X}_{t+1} \leftarrow \mathcal{X}_{t+1} + S\left(\mathcal{X}_t, \mathcal{P}^{(t-1)}(\mathbf{z}), \Psi(t)\right)$

\tcc{\textcolor{blue}{Sample from currently pruned set}}
$\mathcal{X}_{t+1} \leftarrow \mathcal{X}_{t+1} + S\left(\Tilde{\mathcal{X}}_{t}, \mathcal{P}^{(t-1)}(\mathbf{z}), \Tilde{\Psi}(t)\right)$

\tcc{\textcolor{blue}{Standard GNN training}}
Compute loss $\mathcal{L}_{\ourmethod} = \mathcal{L}_{\text{task}} + \lambda_1\cdot\mathcal{L}_{\text{comp}} + \lambda_2\cdot\mathcal{L}_{\text{sepa}}$\Comment*[r]{\textcolor{blue}{Eq.~\ref{eq:loss}}}

Backpropagate to update the GNN model $f_\theta$, projector $g_\phi$, and prototypes.
}

\end{algorithm}

\section{Extension of \ourmethod}\label{app:extend}

In this section, we will explain how to extend \ourmethod beyond traditional graph classification tasks to more complex scenarios like graph regression and graph pre-training. As noted in \Cref{sec:opt}, tasks such as graph pre-training do not have ground truth class indices, making direct application of \ourmethod, which relies on true class labels in \Cref{eq:loss_gder,eq:sep_loss}, infeasible.

A straightforward approach is to assign \(C\) virtual classes, each with \(K\) prototypes \(\mathbf{P}^c = \{\mathbf{p}_k^c\}_{k=1}^K\). During prototype allocation, we use the probability distribution provided by \Cref{eq:mixture_vmf} to determine the class of each graph sample:
\begin{equation}
\Tilde{y}_i = \operatorname{arg max}\limits_c  p(y_i=c|\mathbf{z}_i, \{\mathbf{P}^j,\kappa\}_{j=1}^C).
\end{equation}
We then substitute \(\Tilde{y}_i\) for \(y_i\) in \Cref{eq:loss_gder,eq:sep_loss}, essentially emphasizing that the sample \(\mathbf{z}_i\) should cluster tightly around its assigned prototype cluster \(\chi_{\Tilde{y}_i}\) and remain distant from other clusters. However, this approach is prone to error accumulation: if a sample is initially misclassified, \(\mathcal{L}_{\text{comp}}\) and \(\mathcal{L}_{\text{sepa}}\) will erroneously encourage it to continue moving in the wrong direction. To address this issue, we draw inspiration from previous practices in prototypical contrastive learning and leverage the {prototypical contrastive loss}~\cite{li2020prototypical}:
\begin{equation}
\label{eqn:proto_NCE}
\mathcal{L}_\mathrm{contra} = \sum_{i=1}^{|\mathcal{X}_t|}  -\bigg(\frac{1}{C}\sum_{c=1}^C \log \frac{\sum_{k=1}^{K}\exp(\mathbf{z}_i\cdot \mathbf{p}_s^c/\phi_s^c)}{\sum_{j=0}^{r}\sum_{k=1}^{K} \exp(\mathbf{z}_i\cdot \mathbf{p}_j^c/\phi_j^m)} \bigg),
\end{equation}
where $\phi$ calculates the concentration level of the feature distribution around a prototype as defined in \cite{li2020prototypical}. \Cref{eqn:proto_NCE} has been shown to learn cluster distributions with high mutual information with ground truth labels in unsupervised settings. It encourages samples to migrate between clusters by measuring a concentration-weighted contrastive signal, rather than accumulating current errors. Thus, the overall objective of \ourmethod becomes:

\begin{equation}
\mathcal{L}'_{\ourmethod} = \mathcal{L}_{\text{task}} + \lambda_1\cdot\mathcal{L}_{\text{comp}} + \lambda_2\cdot\mathcal{L}_{\text{sepa}} + \lambda_3\cdot\mathcal{L}_{\text{contra}}.
\end{equation}

We applied this setting when extending \ourmethod to pre-training with GraphMAE on the ZINC dataset, with the experimental results in \Cref{tab:pretrain}.

\section{Additional Experimental Results}\label{app:exp_additional}

We place additional results about \textsc{MUTAG} and \textsc{DHFR} in \Cref{tab:rq1_small_pna}, and the GraphMAE+ZINC pre-training results in \Cref{tab:pretrain}.

\begin{table*}[tp]
    \caption{Graph pre-training performance of \ourmethod on GraphMAE~\cite{hou2022graphmae}+ZINC15~\cite{sterling2015zinc}. Following \cite{hou2022graphmae}, the model is first
pre-trained in 2 million unlabeled molecules sampled from the
ZINC15, and then finetuned in 3 classification benchmark
datasets contained in MoleculeNet~\cite{wu2018moleculenet}.%\footnote{The original pre-training time is $4.8$ h.}
    }
\label{tab:pretrain}
    % \centering
    % \footnotesize
    \setlength{\tabcolsep}{3pt}
    % \begin{adjustbox}{max_width=\textwidth}
    \resizebox{\textwidth}{!}{
    \begin{tabular}{cc|ccc|ccc|ccc}
    \toprule
    % \multicolumn{4}{c}{\textsc{mutag}}  \\
    % \midrule
    & Remaining Ratio \% & \multicolumn{3}{c|}{30\%} & \multicolumn{3}{c|}{50\%}  & \multicolumn{3}{c}{70\%}  \\ \midrule

    & Dataset & BBBP & ToxCast & BACE  & BBBP & ToxCast & BACE  & BBBP & ToxCast & BACE\\ \midrule
    & Original 
    & 72.04 & 65.77 & 81.96 & 72.04 & 65.77 & 81.96  & 72.04 & 65.77 & 81.96   \\

    & +\ourmethod
    & 73.57 & 63.55 & 78.42 & 73.99 & 64.16 & 82.29 & 73.87 & 65.68 &  82.70  \\

    \midrule

     & Time consumption & \multicolumn{3}{c|}{$1.70$ h} & \multicolumn{3}{c|}{$2.58$ h} & \multicolumn{3}{c}{$3.78$ h}\\
    & {Training Speedup} & \multicolumn{3}{c|}{$2.81\times$} & \multicolumn{3}{c|}{$1.86\times$} & \multicolumn{3}{c}{$1.26\times$}\\ 
    \bottomrule
    \end{tabular}
    }
    \footnotesize{The original pre-training time is $4.8$ h.}\\
    % \end{adjustbox}
\end{table*}

\begin{table*}[tp]
    \caption{Performance comparison to state-of-the-art dataset pruning methods. All methods are trained using \textbf{PNA}, and the reported metrics represent the average of \textbf{twenty random runs} and different dataset splits. 
    }
\label{tab:rq1_small_pna}
    \centering
    \footnotesize
    \setlength{\tabcolsep}{3pt}
    % \begin{adjustbox}{max_width=\textwidth}
    \resizebox{\textwidth}{!}{
    \begin{tabular}{cc|cccc|cccc}
    \toprule
     \multirow{3}{*}{} & Dataset   & \multicolumn{4}{c|}{\textsc{mutag} (Accuracy $\uparrow$)} & \multicolumn{4}{c}{\textsc{dhfr} (Accuracy $\uparrow$)}  \\
    \midrule
    & Remaining Ratio \% & 20 & 30 & 50 & 70  & 20& 30 & 50 & 70 \\ \midrule
    \parbox[t]{4mm}{\multirow{17}{*}{\rotatebox[origin=c]{90}{Static}}}
    & Random
    & 85.3\blue{4.1} & 85.6\blue{3.8} & 86.7\blue{2.7} & 88.3\blue{1.1} & 72.3\blue{4.2} & 72.6\blue{3.9} & 73.7\blue{2.8} & 75.8\blue{0.7} \\
    & CD~\cite{agarwal2020contextual}
    & 85.1\blue{4.3} & 85.8\blue{3.6} & 87.0\blue{2.4} & 88.1\blue{1.3}  & 72.1\blue{4.4} & 72.8\blue{3.7} & 74.0\blue{2.5} & 76.1\blue{0.4}\\
    & Herding~\cite{welling2009herding}
    & 77.7\blue{11.7} & 79.6\blue{9.8} & 81.5\blue{7.9} & 87.9\blue{1.5}  & 65.8\blue{10.7} & 67.6\blue{8.9} & 69.6\blue{6.9} & 73.6\blue{2.9}\\
    & K-Center~\cite{sener2018active}
    & 76.2\blue{3.2} & 80.6\blue{8.8} & 84.2\blue{5.2} & 88.4\blue{1.0}  & 64.2\blue{12.3} & 68.2\blue{8.3} & 70.6\blue{5.9}& 72.8\blue{3.7}\\
    & Least Confidence~\cite{coleman2019selection}
    & 85.3\blue{4.1} & 85.6\blue{3.8} & 87.8\blue{1.6} & 88.3\blue{1.1} & 72.3\blue{4.2} & 72.6\blue{3.9} & 74.8\blue{1.7}& 76.1\blue{0.4}\\
    & Margin~\cite{coleman2019selection}
    & 84.2\blue{5.2} & 84.5\blue{4.9} & 87.3\blue{2.1} & 88.4\blue{1.0}& 70.2\blue{6.3} & 71.5\blue{5.0} & 74.6\blue{1.9} & 75.6\blue{0.9}\\
    & Forgetting~\cite{toneva2018empirical}
    & 85.8\blue{3.6} & 86.2\blue{3.2} & 87.1\blue{2.3} & 88.4\blue{1.0}& 72.8\blue{3.7} & 73.2\blue{3.3} & 74.1\blue{2.4}& 76.0\blue{0.5}\\
    & GraNd-4~\cite{paul2021deep}
    & 81.7\blue{7.7} & 85.9\blue{3.5} & 87.0\blue{2.4} & 88.2\blue{1.2} & 68.7\blue{7.8} & 72.9\blue{3.6} & 74.0\blue{2.5} & 75.6\blue{0.9}\\
    & DeepFool~\cite{ducoffe2018adversarial}
    & 85.1\blue{4.3} & 85.6\blue{3.8} & 86.7\blue{2.7} & 88.1\blue{1.3} & 72.1\blue{4.4} & 72.7\blue{3.8} & 73.2\blue{3.3} & 75.8\blue{0.7}\\
    & Craig~\cite{mirzasoleiman2020coresets}
    & 85.0\blue{4.4} & 85.4\blue{4.0} & 86.3\blue{3.1} & 88.2\blue{1.2} & 72.0\blue{4.5} & 72.5\blue{4.0} & 73.7\blue{2.8}& 76.2\blue{0.3}\\
    & Glister~\cite{killamsetty2021glister}
    & 86.3\blue{3.1} & 86.8\blue{2.6} & 87.2\blue{2.2} & 88.4\blue{1.0}& 72.9\blue{3.6} & 73.3\blue{3.2} & 75.2\blue{1.3}& 76.4\blue{0.1}\\
    & Influence~\cite{koh2017understanding}
    & 84.7\blue{4.7} & 85.9\blue{3.5} & 86.7\blue{2.7} & 88.1\blue{1.3} & 71.7\blue{4.8} & 72.9\blue{3.6} & 73.7\blue{2.8} & 75.4\blue{1.1}\\
    & EL2N-2~\cite{toneva2018empirical}
    & 86.2\blue{3.2} & 87.1\blue{2.3} & 87.7\blue{1.7} & 88.2\blue{1.2} & 72.2\blue{4.3} & 73.9\blue{2.6} & 74.6\blue{1.9} & 75.0\blue{1.5}\\
    & EL2N-20~\cite{toneva2018empirical}
    & 86.3\blue{3.1} & \textbf{87.1\blue{2.3}} & 87.9\blue{1.5} & 88.3\blue{1.1}& 72.4\blue{4.1} & 73.0\blue{3.5} & 74.6\blue{1.9}& 76.2\blue{0.3}\\
    & DP~\cite{yang2023dataset}
    & 85.3\blue{4.1} & 86.2\blue{3.2} & 87.4\blue{2.0} & 87.9\blue{1.5} & 71.3\blue{5.2} & 72.6\blue{3.9} & 74.0\blue{2.5} & 75.6\blue{0.9}\\
    \midrule
    \parbox[t]{4mm}{\multirow{4}{*}{\rotatebox[origin=c]{90}{Dynamic}}}
    & Random*
    & 87.0\blue{2.4} & 86.6\blue{2.8} & 88.8\blue{0.6} & 88.9\blue{0.5} & 73.4\blue{3.1}  & 73.9\blue{2.6}  & 74.8\blue{1.7}  & 76.4\blue{0.1} \\

    & $\epsilon$-greedy~\cite{raju2021ddp}
    & 86.5\blue{2.9} & 86.3\blue{3.1} & 88.3\blue{1.1} & 88.9\blue{0.5}& 73.1\blue{3.4} & 73.4\blue{3.1} & 74.1\blue{1.6}& 76.4\blue{0.1}\\

    & UCB~\cite{raju2021ddp}
    & 86.5\blue{2.9} & 86.4\blue{3.0} & 87.7\blue{1.7} & 88.5\blue{0.9}& 73.1\blue{3.4} & 73.5\blue{3.0} & 74.0\blue{2.5}& 75.9\blue{0.6}\\

    & InfoBatch~\cite{qin2023infobatch}
    & 86.8\blue{2.6} & 86.7\blue{2.7} & 89.0\blue{0.4} & 89.3\blue{0.1}& 73.3\blue{3.2} & 73.7\blue{2.8} & 75.0\blue{1.5} & 76.3\blue{0.2}\\

    & \ourmethod 
    & \colorbox[HTML]{DAE8FC}{\textbf{88.2\blue{\textbf{1.2}}}} 
    & \colorbox[HTML]{DAE8FC}{\textbf{88.5\blue{\textbf{0.9}}} }
    &  \colorbox[HTML]{DAE8FC}{\textbf{89.3\blue{\textbf{0.1}}}} 
    & \colorbox[HTML]{DAE8FC}{\textbf{89.9\red{\textbf{0.5}}}}
    & \colorbox[HTML]{DAE8FC}{\textbf{75.7\blue{\textbf{0.8}}} }
    & \colorbox[HTML]{DAE8FC}{\textbf{75.9\blue{\textbf{0.6}}}}
    & \colorbox[HTML]{DAE8FC}{\textbf{76.1\blue{\textbf{0.4}}} }
    & \colorbox[HTML]{DAE8FC}{\textbf{77.1\red{\textbf{0.6}}}} \\

    \midrule
    \multicolumn{2}{c|}{Whole Dataset} & \multicolumn{4}{c|}{89.4$_{\pm0.1}$} & \multicolumn{4}{c}{76.5$_{\pm0.1}$} \\
    \bottomrule
    \end{tabular}}
    % \end{adjustbox}
\end{table*}

\section{Supplementary Related Work}

\paragraph{Constrastive Learning and Prototypical learning}  Contrastive representation learning methods consider each sample as a unique class, aligning multiple views of the same input while distancing other samples. This significantly improves the discriminative power of the learned representations, allowing these methods to excel in learning robust feature representations across unsupervised \cite{wu2018unsupervised,oord2018representation,he2020momentum,chen2020simple,robinson2021contrastive}, semi-supervised \cite{assran2021semi}, and supervised settings \cite{khosla2020supervised}. The foundational properties and effectiveness of contrastive loss within hyperspherical space have been extensively studied \cite{wang2020understanding,wang2021understanding}. Other approaches focus on learning feature representations by modeling the relationships between samples and cluster centroids \cite{caron2018deep} or prototypes \cite{snell2017prototypical}. Building on contrastive learning, \cite{li2021prototypical} incorporates prototypical learning, adding a contrastive mechanism between samples and prototypes obtained through offline clustering. 
PALM~\cite{lu2024learning} utilizes prorotypical learning for out-of-distribution (OOD) identification, which automatically identifies and dynamically updates prototypes, and assigns each sample to a subset of prototypes via reciprocal neighbor soft assignment weights. However, all these methods are not conducive to a more lightweight training burden, and our method is the first attempt at leveraging prototype learning for soft data pruning.

\end{appendices}

% Optionally include supplemental material (complete proofs, additional experiments and plots) in appendix.
% All such materials \textbf{SHOULD be included in the main submission.}

%%%%%%%%%%%%%%%%%%%%%%%%%%%%%%%%%%%%%%%%%%%%%%%%%%%%%%%%%%%%

\newpage
\section*{NeurIPS Paper Checklist}

\begin{enumerate}

\item {\bf Claims}
    \item[] Question: Do the main claims made in the abstract and introduction accurately reflect the paper's contributions and scope?
    \item[] Answer: \answerYes{} % Replace by \answerYes{}, \answerNo{}, or \answerNA{}.
    \item[] Justification: In this paper, we introduce a novel soft pruning strategy and we claim the contributions and scope in the abstract and introduction sections (See Abstract and Introduction Section).

\item {\bf Limitations}
    \item[] Question: Does the paper discuss the limitations of the work performed by the authors?
    \item[] Answer: \answerYes{} % Replace by \answerYes{}, \answerNo{}, or \answerNA{}.
    \item[] Justification: In this work, we systematically discuss the limitations of our research and outline directions for future work (See Introduction Section).

\item {\bf Theory Assumptions and Proofs}
    \item[] Question: For each theoretical result, does the paper provide the full set of assumptions and a complete (and correct) proof?
    \item[] Answer: \answerNA{} % Replace by \answerYes{}, \answerNo{}, or \answerNA{}.
    \item[] Justification: This paper does not include experimental results related to theoretical aspects.

    \item {\bf Experimental Result Reproducibility}
    \item[] Question: Does the paper fully disclose all the information needed to reproduce the main experimental results of the paper to the extent that it affects the main claims and/or conclusions of the paper (regardless of whether the code and data are provided or not)?
    \item[] Answer: \answerYes{} % Replace by \answerYes{}, \answerNo{}, or \answerNA{}.
    \item[] Justification: We provide the code necessary for replicating the studies described in this paper via an anonymous link, and we detail the experimental setup for the replication in the article itself (See Appendix).

\item {\bf Open access to data and code}
    \item[] Question: Does the paper provide open access to the data and code, with sufficient instructions to faithfully reproduce the main experimental results, as described in supplemental material?
    \item[] Answer: \answerYes{} % Replace by \answerYes{}, \answerNo{}, or \answerNA{}.
    \item[] Justification: For the datasets disclosed in the article, we have provided information regarding their sources and origins (See Appendix).

\item {\bf Experimental Setting/Details}
    \item[] Question: Does the paper specify all the training and test details (e.g., data splits, hyperparameters, how they were chosen, type of optimizer, etc.) necessary to understand the results?
    \item[] Answer: \answerYes{} % Replace by \answerYes{}, \answerNo{}, or \answerNA{}.
    \item[] Justification: we have specified all the training and test details (e.g., data splits, hyperparameters, how they were chosen, type of optimizer, etc.) necessary to understand the results (See Appendix B).

\item {\bf Experiment Statistical Significance}
    \item[] Question: Does the paper report error bars suitably and correctly defined or other appropriate information about the statistical significance of the experiments?
    \item[] Answer: \answerYes{} % Replace by \answerYes{}, \answerNo{}, or \answerNA{}.
    \item[] Justification: In this paper, we have reported the standard deviation of the experiments (See Experiments).

\item {\bf Experiments Compute Resources}
    \item[] Question: For each experiment, does the paper provide sufficient information on the computer resources (type of compute workers, memory, time of execution) needed to reproduce the experiments?
    \item[] Answer: \answerYes{} % Replace by \answerYes{}, \answerNo{}, or \answerNA{}.
    \item[] Justification: In this paper, we provide detailed information about the experimental resources, including GPU configurations used in our studies (See Appendix). 
    
\item {\bf Code Of Ethics}
    \item[] Question: Does the research conducted in the paper conform, in every respect, with the NeurIPS Code of Ethics \url{https://neurips.cc/public/EthicsGuidelines}?
    \item[] Answer: \answerYes{} % Replace by \answerYes{}, \answerNo{}, or \answerNA{}.
    \item[] Justification: The study presented in this paper conforms to the NeurIPS Code of Ethics.

\item {\bf Broader Impacts}
    \item[] Question: Does the paper discuss both potential positive societal impacts and negative societal impacts of the work performed?
    \item[] Answer: \answerYes{} % Replace by \answerYes{}, \answerNo{}, or \answerNA{}.
    \item[] Justification: We have provided the societal impacts of the work (See Introduction).
    
\item {\bf Safeguards}
    \item[] Question: Does the paper describe safeguards that have been put in place for responsible release of data or models that have a high risk for misuse (e.g., pretrained language models, image generators, or scraped datasets)?
    \item[] Answer: \answerNA{} % Replace by \answerYes{}, \answerNo{}, or \answerNA{}.
    \item[] Justification: This paper does not address issues related to this aspect.

\item {\bf Licenses for existing assets}
    \item[] Question: Are the creators or original owners of assets (e.g., code, data, models), used in the paper, properly credited and are the license and terms of use explicitly mentioned and properly respected?
    \item[] Answer: \answerYes{} % Replace by \answerYes{}, \answerNo{}, or \answerNA{}.
    \item[] Justification: All creators and original owners of the assets used in our paper, such as code, data, and models, have been properly credited. We have explicitly mentioned the licenses and terms of use for each asset and have ensured full compliance with these terms throughout our research.

\item {\bf New Assets}
    \item[] Question: Are new assets introduced in the paper well documented and is the documentation provided alongside the assets?
    \item[] Answer: \answerNA{} % Replace by \answerYes{}, \answerNo{}, or \answerNA{}.
    \item[] Justification: The research presented in this paper is not concerned with new assets.

\item {\bf Crowdsourcing and Research with Human Subjects}
    \item[] Question: For crowdsourcing experiments and research with human subjects, does the paper include the full text of instructions given to participants and screenshots, if applicable, as well as details about compensation (if any)? 
    \item[] Answer: \answerNA{} % Replace by \answerYes{}, \answerNo{}, or \answerNA{}.
    \item[] Justification: This paper does not involve experiments or research related to human subjects.

\item {\bf Institutional Review Board (IRB) Approvals or Equivalent for Research with Human Subjects}
    \item[] Question: Does the paper describe potential risks incurred by study participants, whether such risks were disclosed to the subjects, and whether Institutional Review Board (IRB) approvals (or an equivalent approval/review based on the requirements of your country or institution) were obtained?
    \item[] Answer: \answerNA{} % Replace by \answerYes{}, \answerNo{}, or \answerNA{}.
    \item[] Justification: This paper does not address potential risks incurred by study participants.

\end{enumerate}

\end{document}